\documentclass[runningheads]{llncs}
\usepackage{graphicx}

\usepackage[pagebackref,breaklinks,colorlinks]{hyperref}

\usepackage{tikz}
\usepackage{comment}
\usepackage{amsmath,amssymb} 
\usepackage{graphicx}
\usepackage[caption=false]{subfig}
\usepackage{booktabs}
\usepackage{color, colortbl}
\usepackage{wrapfig}
\usepackage{cite}
\usepackage{xspace}
\usepackage[symbol]{footmisc}
\usepackage[style=plain]{floatrow}
\usepackage{capt-of}

\usepackage[accsupp]{axessibility}  

\definecolor{LightYellow}{rgb}{0.99,0.94,0.7}

\newcommand{\ignore}[1]{}

\DeclareMathOperator{\softmax}{Softmax}
\DeclareMathOperator{\cdf}{CDF}
\newcommand{\expnumber}[2]{{#1}\mathrm{e}{#2}}
\newcommand{\round}[1]{\ensuremath{\lfloor#1\rceil}}

\makeatletter
\DeclareRobustCommand\onedot{\futurelet\@let@token\@onedot}
\def\@onedot{\ifx\@let@token.\else.\null\fi\xspace}

\def\eg{\emph{e.g}\onedot} 
\def\ie{\emph{i.e}\onedot}

\makeatother

\begin{document}
\pagestyle{headings}
\mainmatter

\def\ECCVSubNumber{4901}  

\title{Adaptive Token Sampling For Efficient Vision Transformers} 

\titlerunning{Adaptive Token Sampling}
%
\renewcommand{\thefootnote}{\fnsymbol{footnote}}
\author{Mohsen Fayyaz$^{1, 6, } \thanks{Equal Contribution} \thanks{Work has been done during an internship at Microsoft} \quad$ 
Soroush Abbasi Koohpayegani$^{2, }{^*}{^\dagger}\quad$ 
Farnoush Rezaei Jafari$^{3,4, }{^*}\quad$
Sunando Sengupta$^{1}\quad$
Hamid Reza Vaezi Joze$^{5}\quad$
Eric Sommerlade$^{1}\quad$
Hamed Pirsiavash$^{2}\quad$
Juergen Gall$^{6}\quad$
}

%
\institute{
$^{1}$Microsoft\quad $^{2}$University of California, Davis \quad $^{3}$Machine Learning Group, Technische Universit\"at Berlin, $^{4}$Berlin Institute for the Foundations of Learning and Data \quad $^{5}$Meta Reality Labs  \quad $^{6}$University of Bonn 
}

\authorrunning{Fayyaz, Abbasi Koohpayegani, Rezaei Jafari et al.} 
\maketitle

\begin{abstract}

While state-of-the-art vision transformer models achieve promising results in image classification, they are computationally expensive and require many GFLOPs. Although the GFLOPs of a vision transformer can be decreased by reducing the number of tokens in the network, there is no setting that is optimal for all input images. In this work, we therefore introduce a differentiable parameter-free Adaptive Token Sampler (ATS) module, which can be plugged into any existing vision transformer architecture. ATS empowers vision transformers by scoring and adaptively sampling significant tokens. As a result, the number of tokens is not constant anymore and varies for each input image. By integrating ATS as an additional layer within the current transformer blocks, we can convert them into much more efficient vision transformers with an adaptive number of tokens. Since ATS is a parameter-free module, it can be added to the off-the-shelf pre-trained vision transformers as a plug and play module, thus reducing their GFLOPs without any additional training. Moreover, due to its differentiable design, one can also train a vision transformer equipped with ATS. We evaluate the efficiency of our module in both image and video classification tasks by adding it to multiple SOTA vision transformers.
Our proposed module improves the SOTA by reducing their computational costs (GFLOPs) by 2$\times$, while preserving their accuracy on the ImageNet, Kinetics-400, and Kinetics-600 datasets. The code is available at \href{https://adaptivetokensampling.github.io/}{https://adaptivetokensampling.github.io/}.

\end{abstract}

\section{Introduction}
\begin{figure*}[t!]
\begin{center}
\end{center}
   \includegraphics[width=1.0\linewidth]{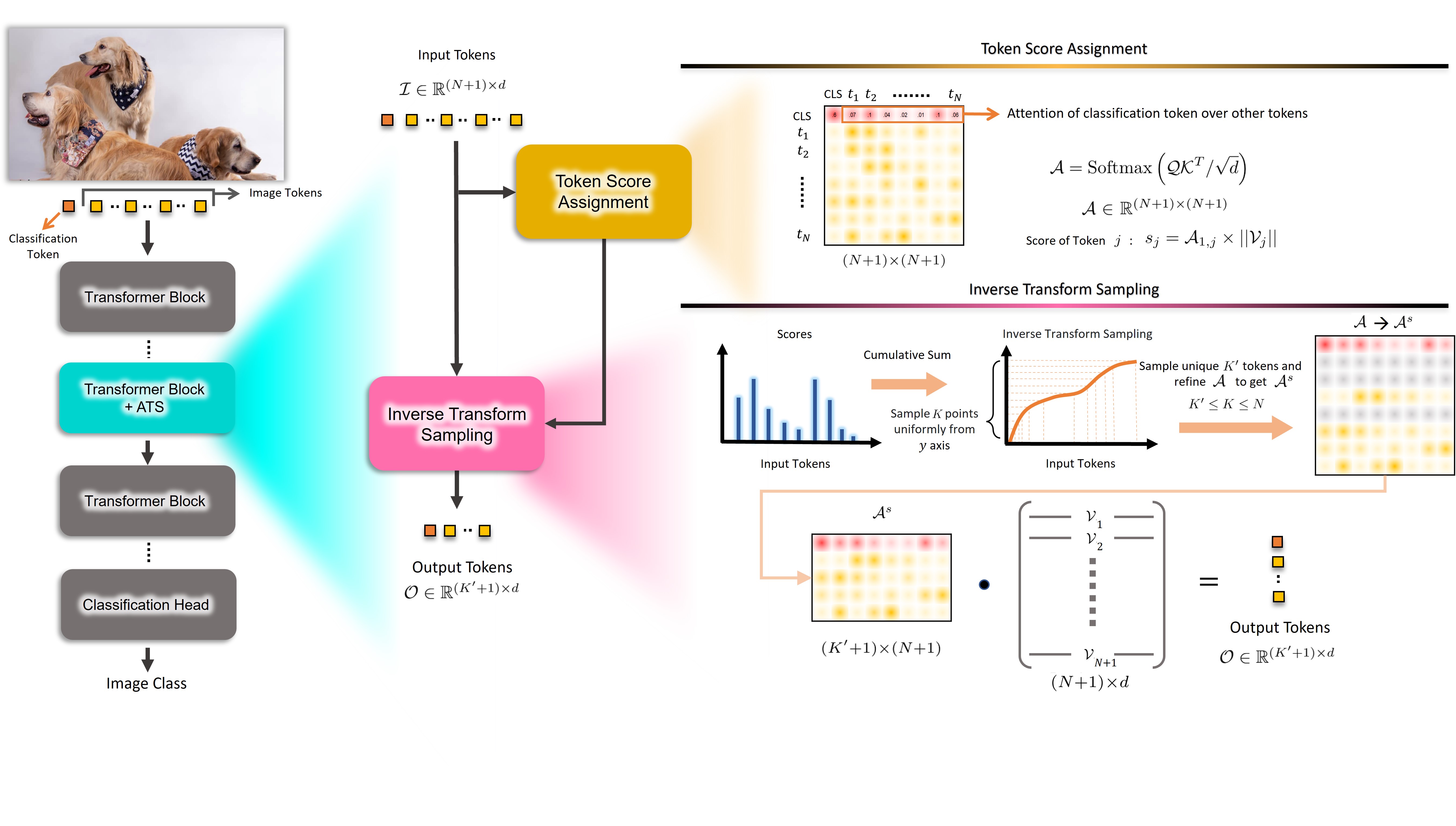}
   \caption{The Adaptive Token Sampler (ATS) can be integrated into the self-attention layer of any transformer block of a vision transformer model (left). 
   The ATS module takes at each stage a set of input tokens $\mathcal{I}$. The first token is considered as the classification token in each block of the vision transformer. The attention matrix $\mathcal{A}$ is then calculated by the dot product of the queries $\mathcal{Q}$ and keys $\mathcal{K}$, scaled by $\sqrt{d}$. We use the attention weights $\mathcal{A}_{1,2}, \ldots, \mathcal{A}_{1,N+1}$ of the classification token as significance scores $\mathcal{S}\in \mathbb{R}^{N}$ for pruning the attention matrix $\mathcal{A}$. To reflect the effect of values $\mathcal{V}$ on the output tokens $\mathcal{O}$, we multiply the $\mathcal{A}_{1,j}$ by the magnitude of the corresponding value $\mathcal{V}_j$. We select the significant tokens using inverse transform sampling over the cumulative distribution function of the scores $\mathcal{S}$. Having selected the significant tokens, we then sample the corresponding attention weights (rows of the attention matrix $\mathcal{A}$) to get $\mathcal{A}^s$. Finally, we softly downsample the input tokens $\mathcal{I}$ to output tokens $\mathcal{O}$ using the dot product of $\mathcal{A}^s$ and $\mathcal{V}$. \vspace{-20px}
  }
\label{fig:teaser}
\end{figure*}

Over the last ten years, there has been a tremendous progress on image and video understanding in the light of new and complex deep learning architectures, which are based on the variants of 2D~\cite{he2015deep, simonyan2015deep, krizhevsky2014weird} and 3D~\cite{c3d, feichtenhofer2019slowfast, diba2018spatio,dynamonet, res3d, feichtenhofer2020x3d} Convolutional Neural Networks (CNNs). Recently, vision transformers have shown promising results in image classification \cite{vit, deit, token_labeling, cvt} and action recognition \cite{bertasius2021space, swin, xvit} compared to CNNs. Although vision transformers have a superior representation power, the high computational cost of their transformer blocks make them unsuitable for many edge devices.
The computational cost of a vision transformer grows quadratically with respect to the number of tokens it uses. To reduce the number of tokens and thus the computational cost of a vision transformer, DynamicViT \cite{dynamicvit} proposes a token scoring neural network to predict which tokens are redundant. The approach then keeps a fixed ratio of tokens at each stage. Although DynamicViT reduces the GFLOPs of a given network, its scoring network introduces an additional computational overhead. Furthermore, the scoring network needs to be trained together with the vision transformer and it requires to modify the loss function by adding additional loss terms and hyper-parameters. To alleviate such limitations, EViT\cite{evit} employs the attention weights as the tokens' importance scores. A further limitation of both EViT and DynamicViT is that they need to be re-trained if the fixed target ratios need to be changed (\eg due to deployment on a different device). This strongly limits their applications.            

In this work, we propose a method to efficiently reduce the number of tokens in any given vision transformer without the mentioned limitations. Our approach is motivated by the observation that in image/action classification, all parts of an input image/video do not contribute equally to the final classification scores and some parts contain irrelevant or redundant information. The amount of relevant information varies depending on the content of an image or video. For instance, in Fig.~\ref{fig:dynamicity_visualization}, we can observe examples in which only a few or many patches are required for correct classification. The same holds for the number of tokens used at each stage, as illustrated in Fig.~\ref{fig:token_removal}. Therefore, we propose an approach that automatically selects an adequate number of tokens at each stage based on the image content, \ie the number of the selected tokens at all network's stages varies for different images, as shown in Fig.~\ref{fig:dynamicity}. It is in contrast to \cite{dynamicvit, evit}, where the ratio of the selected tokens needs to be specified for each stage and is constant after training. 
However, selecting a static number of tokens will on the one hand discard important information for challenging images/videos, which leads to a classification accuracy drop. On the other hand, it will use more tokens than necessary for the easy cases and thus waste computational resources. 
In this work, we address the question of how a transformer can dynamically adapt its computational resources in a way that not more resources than necessary are used for each input image/video.

To this end, we introduce a novel \emph{Adaptive Token Sampler (ATS)} module. ATS is a differentiable parameter-free module that adaptively down-samples input tokens. To do so, we first assign significance scores to the input tokens by employing the attention weights of the classification token in the self-attention layer and then select a subset of tokens using inverse transform sampling over the scores. Finally, we softly down-sample the output tokens to remove redundant information with the least amount of information loss. In contrast to \cite{dynamicvit}, our approach does not add any additional learnable parameters to the network. While the ATS module can be added to any off-the-shelf pre-trained vision transformer without any further training, the network equipped with the differentiable ATS module can also be further fine-tuned. 
Moreover, one may train a model only once and then adjust a maximum limit for the ATS module to adapt it to the resources of different edge devices at the inference time. This eliminates the need of training separate models for different levels of computational resources.

We demonstrate the efficiency of our proposed adaptive token sampler for image classification by integrating it into the current state-of-the-art vision transformers such as DeiT~\cite{deit}, CvT~\cite{cvt}, and PS-ViT~\cite{psvit}.
As shown in Fig.~\ref{fig:model_scaling}, our approach significantly reduces the GFLOPs of vision transformers of various sizes without significant loss of accuracy. We evaluate the effectiveness of our method by comparing it with other methods designed for reducing the number of tokens, including DynamicViT~\cite{dynamicvit}, EViT~\cite{evit}, and Hierarchical Pooling~\cite{HVT}. Extensive experiments on the ImageNet dataset show that our method outperforms existing approaches and provides the best trade-off between computational cost and classification accuracy. We also demonstrate the efficiency of our proposed module for action recognition by adding it to the state-of-the-art video vision transformers such as XViT~\cite{xvit} and TimeSformer~\cite{bertasius2021space}. Extensive experiments on the Kinetics-400 and Kinetics-600 datasets show that our method surpasses the performance of existing approaches and leads to the best computational cost/accuracy trade-off.
In a nutshell, the adaptive token sampler can significantly scale down the off-the-shelf vision transformers' computational costs and it is therefore very useful for real-world vision-based applications.  

\section{Related Work}
The transformer architecture, which was initially introduced in the NLP community \cite{attention_is_all_you_need}, has demonstrated promising performance on various computer vision tasks \cite{vit, deit, swin, deepvit, global_filter_net, detr, setr, maskformer, pointr, pointtransformer}. ViT \cite{vit} follows the standard transformer architecture to tailor a network that is applicable to images. It splits an input image into a set of non-overlapping patches and produces patch embeddings of lower dimensionality. The network then adds positional embeddings to the patch embeddings and passes them through a number of transformer blocks. An extra learnable class embedding is also added to the patch embeddings to perform classification. Although ViT has shown promising results in image classification, it requires an extensive amount of data to generalize well. DeiT \cite{deit} addressed this issue by introducing a distillation token designed to learn from a teacher network. Additionally, it surpassed the performance of ViT. LV-ViT \cite{token_labeling} proposed a new objective function for training vision transformers and achieved better performance. TimeSformer \cite{bertasius2021space} proposed a new architecture for video understanding by extending the self-attention mechanism of the standard transformer models to video. The complexity of the TimeSformer's self-attention is $O(T^2S+TS^2)$ where $T$ and $S$ represent temporal and spatial locations respectively. X-ViT \cite{xvit} reduced this complexity to $O(TS^2)$ by proposing an efficient video transformer.

Besides the accuracy of neural networks, their efficiency plays an important role in deploying them on edge devices. A wide range of techniques have been proposed to speed up the inference of these models. To obtain deep networks that can be deployed on different edge devices, works like \cite{efficient_net} proposed more efficient architectures by carefully scaling the depth, width, and resolution of a baseline network based on different resource constraints. \cite{mobilenets} aims to meet such resource requirements by introducing hyper-parameters, which can be tuned to build efficient light-weight models. The works \cite{gong2014compressing, haq} have adopted quantization techniques to compress and accelerate deep models. Besides quantization techniques, other approaches such as channel pruning \cite{channel_pruning}, run-time neural pruning \cite{runtime_net_prunning}, low-rank matrix decomposition \cite{low_rank_factorization_1, low_rank_factorization_2}, and knowledge distillation \cite{hinton2015distilling, metadistiller} have been used as well to speed up deep networks.

In addition to the works that aim to accelerate the inference of convolutional neural networks, other works aim to improve the efficiency of transformer-based models. In the NLP area, Star-Transformer \cite{startransformer} reduced the number of connections from \(n^2\) to \(2n\) by changing the fully-connected topology into a star-shaped structure. TinyBERT \cite{tinybert} improved the network's efficiency by distilling the knowledge of a large teacher BERT into a tiny student network. PoWER-BERT \cite{powerbert} reduced the inference time of the BERT model by identifying and removing redundant and less-informative tokens based on their importance scores estimated from the self-attention weights of the transformer blocks. To reduce the number of FLOPs in character-level language modeling, a new self-attention mechanism with adaptive attention span is proposed in \cite{adaptive-span}. To enable fast performance in unbatched decoding and improve the scalability of the standard transformers, Scaling Transformers \cite{scaling-transformers} are introduced. These novel transformer architectures are equipped with sparse variants of standard transformer layers.

To improve the efficiency of vision transformers, sparse factorization of the dense attention matrix has been proposed \cite{sparsetransformer}, which reduces its complexity to $O(n\sqrt{n})$ for the autoregressive image generation task. \cite{routingtransformer} tackled this problem by proposing an approach to sparsify the attention matrix. They first cluster all the keys and queries and only consider the similarities of the keys and queries that belong to the same cluster. DynamicViT \cite{dynamicvit} proposed an additional prediction module that predicts the importance of tokens and discards uninformative tokens for the image classification task. Hierarchical Visual Transformer (HVT) \cite{HVT} employs token pooling, which is similar to feature map down-sampling in convolutional neural networks, to remove redundant tokens. PS-ViT \cite{psvit} incorporates a progressive sampling module that iteratively learns to sample distinctive input tokens instead of uniformly sampling input tokens from all over the image. The sampled tokens are then fed into a vision transformer module with fewer transformer encoder layers compared to ViT. TokenLearner \cite{tokenlearner} introduces a learnable tokenization module that can reduce the computational cost by learning few important tokens conditioned on the input. They have demonstrated that their approach can be applied to both image and video understanding tasks. Token Pooling \cite{token-pooling} down-samples tokens by grouping them into a set of clusters and returning the cluster centers. A concurrent work \cite{evit} introduces a token reorganization method that first identifies top-k important tokens by computing token attentiveness between the tokens and the classification token and then fuses less informative tokens. IA-RED$^2$ \cite{IARED} proposes an interpretability-aware redundancy reduction framework for vision transformers that discards less informative patches in the input data. Most of the mentioned approaches improve the efficiency of vision transformers by introducing architectural changes to the original models or by adding modules that add extra learnable parameters to the networks, while our parameter-free adaptive module can be incorporated into off-the-shelf architectures and reduces their computational complexity without significant accuracy drop and even without requiring any further training.

\section{Adaptive Token Sampler}

State-of-the-art vision transformers are computationally expensive since their computational costs grow quadratically with respect to the number of tokens, which is static at all stages of the network and corresponds to the number of input patches. 
Convolutional neural networks deal with the computational cost by reducing the resolution within the network using various pooling operations. 
It means that the spatial or temporal resolution decreases at the later stages of the network. However, applying such simple strategies, \ie pooling operations with fixed kernels, to vision transformers is not straightforward since the tokens are permutation invariant. Moreover, such static down-sampling approaches are not optimal. On the one hand, a fixed down-sampling method discards important information at some locations of the image or video, like details of the object. On the other hand, it still includes many redundant features that do not contribute to the classification accuracy, for instance, when dealing with an image with a homogeneous background. Therefore, we propose an approach that dynamically adapts the number of tokens at each stage of the network based on the input data such that important information is not discarded and no computational resources are wasted for processing redundant information.

To this end, we propose our novel Adaptive Token Sampler (ATS) module. ATS is a parameter-free differentiable module to sample significant tokens over the input tokens. In our ATS module, we first assign significance scores to the $N$ input tokens and then select a subset of these tokens based on their scores. The upper bound of GFLOPs can be set by defining a maximum limit for the number of tokens sampled, denoted by $K$. Since the sampling procedure can sample some input tokens several times, we only keep one instance of a token. The number of sampled tokens $K'$ is thus usually lower than $K$ and varies among input images or videos (Fig.~\ref{fig:dynamicity}). Fig.~\ref{fig:teaser} gives an overview of our proposed approach. 

\subsection{Token Scoring}  
\label{sec:token_score_assignment}
Let $\mathcal{I}\in \mathbb{R}^{(N+1)\times d}$ be the input tokens of a self-attention layer with $N+1$ tokens. Before forwarding the input tokens through the model, ViT concatenates a classification token to the input tokens. The corresponding output token at the final transformer block is then fed to the classification head to get the class probabilities. Practically, this token is placed as the first token in each block and it is considered as a classification token. 
While we keep the classification token, our goal is to reduce the output tokens $\mathcal{O}\in \mathbb{R}^{(K'+1)\times d}$ such that $K'$ is dynamically adapted based on the input image or video and $K'\leq K\leq N$, where $K$ is a parameter that controls the maximum number of sampled tokens. Fig.~\ref{fig:dynamicity} shows how the number of sampled tokens $K'$ varies for different input data and stages of a network. We first describe how each token is scored.

In a standard self-attention layer \cite{attention_is_all_you_need}, the queries $\mathcal{Q} \in \mathbb{R}^{ (N+1) \times d}$, keys $\mathcal{K} \in \mathbb{R}^{ (N+1) \times d}$, and values $\mathcal{V} \in \mathbb{R}^{ (N+1) \times d}$ are computed from the input tokens $\mathcal{I}\in \mathbb{R}^{(N+1)\times d}$. The attention matrix $\mathcal{A}$ is then calculated by the dot product of the queries and keys, scaled by $\sqrt{d}$:
\begin{equation}
    \label{eq:AttentionWeights}
    \mathcal{A} = \softmax\left(\mathcal{QK}^T / \sqrt{d}\right).
\end{equation}
Due to the $\softmax$ function, each row of $\mathcal{A} \in \mathbb{R}^{ (N+1) \times (N+1)}$ sums up to $1$. The output tokens are then calculated using a combination of the values weighted by the attention weights:    
\begin{equation}
    \label{eq:AttentionMechanism}
    \mathcal{O} = \mathcal{AV}.
\end{equation}
Each row of $\mathcal{A}$ contains the attention weights of an output token. The weights indicate the contributions of all input tokens to the output token. Since $\mathcal{A}_{1,:}$ contains the attention weights of the classification token, $\mathcal{A}_{1,j}$ represents the importance of the input token $j$ for the output classification token. Thus, we use the weights $\mathcal{A}_{1,2}, \ldots, \mathcal{A}_{1,N+1}$ as significance scores for pruning the attention matrix $\mathcal{A}$, as illustrated in Fig.~\ref{fig:teaser}. Note that $\mathcal{A}_{1,1}$ is not used since we keep the classification token. As the output tokens $\mathcal{O}$ depend on both $\mathcal{A}$ and $\mathcal{V}$ \eqref{eq:AttentionMechanism}, we also take into account the norm of $\mathcal{V}_j$ for calculating the $j^{th}$ token's significance score. The motivation is that values having a norm close to zero have a low impact and their corresponding tokens are thus less significant. In our experiments, we show that multiplying $\mathcal{A}_{1,j}$ with the norm of $\mathcal{V}_j$ improves the results. The significance score of a token $j$ is thus given by 
\begin{equation}
    \label{eq:score}
    \mathcal{S}_j = \frac{\mathcal{A}_{1,j}\times ||\mathcal{V}_{j}||}{\sum_{i=2} \mathcal{A}_{1,i}\times ||\mathcal{V}_{i}||}
\end{equation}

where $i,j\in \{ 2\dots N\}$. For a multi-head attention layer, we calculate the scores for each head and then sum the scores over all heads.

\subsection{Token Sampling}
\label{sec:score_based_token_sampling}
\begin{figure*}[!htb]
\begin{center}
   \includegraphics[width=0.9\linewidth]{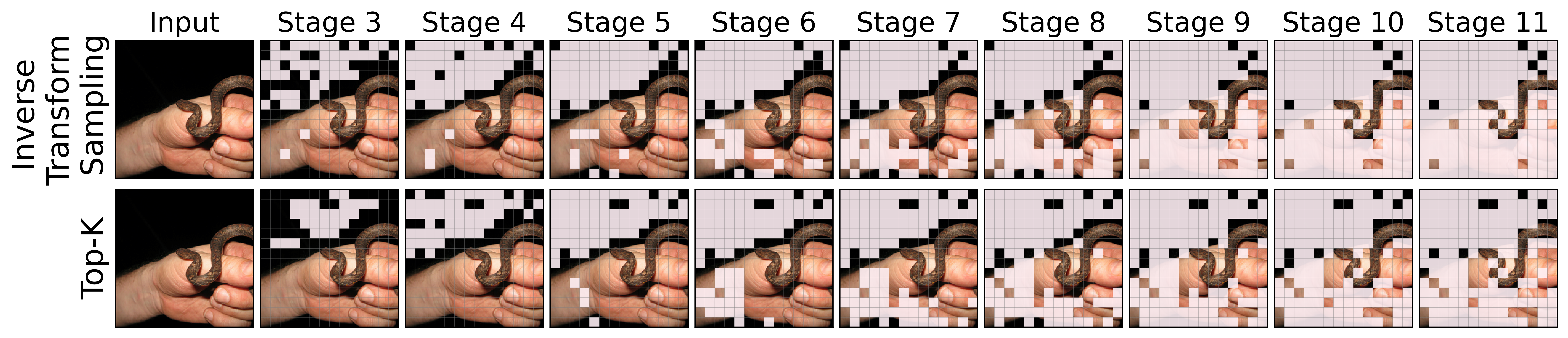}
   \includegraphics[width=0.9\linewidth]{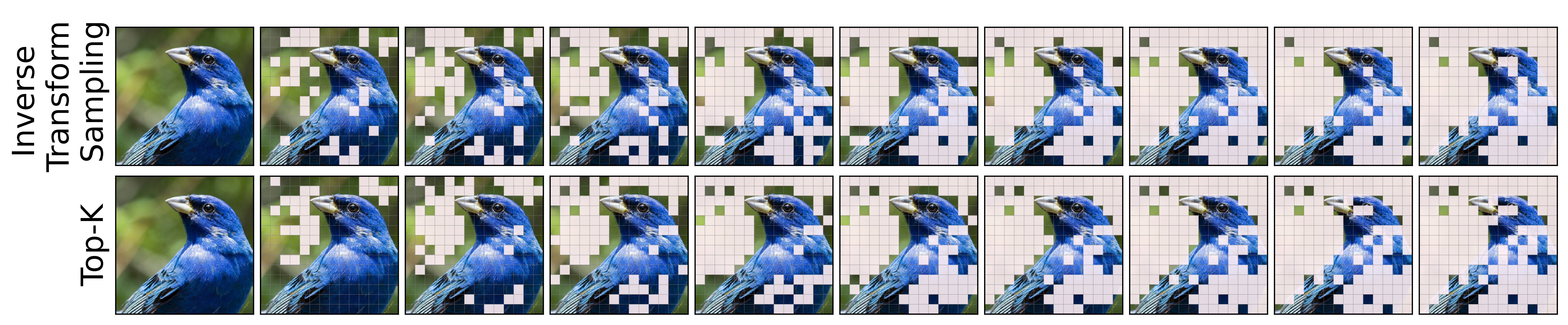}
   \includegraphics[width=0.9\linewidth]{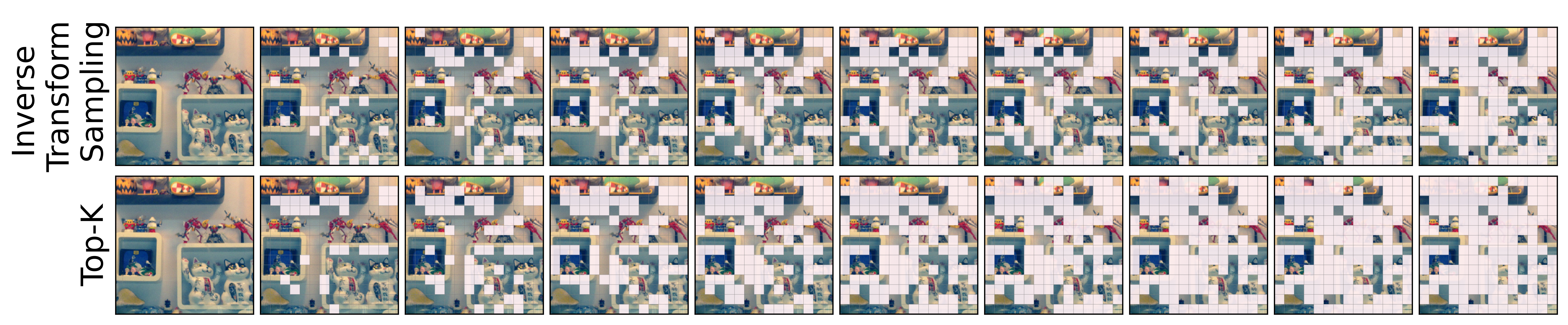}
   \caption{Visualization of the gradual token sampling procedure in the multi-stage DeiT-S+ATS model.
   As it can be seen, at each stage, those tokens that are considered to be less significant to the classification are masked and the ones that have contributed the most to the model's prediction are sampled. We also visualize the token sampling results with Top-K selection to have a better comparison to our Inverse Transform Sampling. }
\label{fig:token_removal}
\end{center}
\end{figure*}

Having computed the significance scores of all tokens, we can prune their corresponding rows from the attention matrix $\mathcal{A}$. To do so, a naive approach is to select $K$ tokens with the highest significance scores (top-$K$ selection). However, this approach does not perform well, as we show in our experiments and it can not adaptively select $K' \leq K$ tokens.
is that it discards all tokens with lower scores. Some of these tokens, however, can be useful in particular at the earlier stages when the features are less discriminative. For instance, having multiple tokens with similar keys, which may occur in the early stages, will lower their corresponding attention weights due to the $\softmax$ function. Although one of these tokens would be beneficial at the later stages, taking the top-$K$ tokens might discard all of them. Therefore, we suggest sampling tokens based on their significance scores. In this case, the probability of sampling one of the several similar tokens is equal to the sum of their scores. We also observe that the proposed sampling procedure selects more tokens at the earlier stages than the later stages as shown in Fig.~\ref{fig:token_removal}.             

For the sampling step, we suggest using inverse transform sampling to sample tokens based on their significance scores $\mathcal{S}$ \eqref{eq:score}. Since the scores are normalized, they can be interpreted as probabilities and we can calculate the cumulative distribution function ($\cdf$) of $\mathcal{S}$: 
\vspace{-.05in}
\begin{equation}
    \label{eq:CDF}
    \cdf_i=\sum_{j=2}^{j=i}\mathcal{S}_j.
\end{equation}
Note that we start with the second token since we keep the first token. 
Having the cumulative distribution function, we obtain the sampling function by taking the inverse of the $\cdf$:   
\vspace{-.05in}
\begin{equation}
    \label{eq:Sampling}
    \Psi(k)=\cdf^{-1}(k)
\end{equation}
where $k \in [0, 1]$. In other words, the significance scores are used to calculate the mapping function between the indices of the original tokens and the sampled tokens. To obtain $K$ samples, we can sample $K$-times from the uniform distribution $U[0, 1]$. While such randomization might be desirable for some applications, deterministic inference is in most cases preferred. Therefore, we use a fixed sampling scheme for training and inference by choosing $k = \{\frac{1}{2K}, \frac{3}{2K} \dots, \frac{2K-1}{2K}\}$. Since $\Psi(.)\in \mathbb{R}$, we consider the indices of the tokens with the nearest significant scores as the sampling indices.

If a token is sampled more than once, we only keep one instance. As a consequence, the number of unique indices $K'$ is often lower than $K$ as shown in Fig.~\ref{fig:dynamicity}. In fact, $K' < K$ if there is at least one token with a score $S_j \geq 2/K$. In the two extreme cases, either only one dominant token is selected and $K'=1$ or $K'=K$ if the scores are more or less balanced. Interestingly, more tokens are selected at the earlier stages, where the features are less discriminative and the attention weights are more balanced, and less at the later stages, as shown in Fig.~\ref{fig:token_removal}. The number and locations of tokens also vary for different input images, as shown in Fig.~\ref{fig:dynamicity_visualization}. For images with a homogeneous background that covers a large part of the image, only a few tokens are sampled. In this case, the tokens cover the object in the foreground and are sparsely but uniformly sampled from the background. In cluttered images, many tokens are required. It illustrates the importance of making the token sampling procedure adaptive.                                  

Having indices of the sampled tokens, we refine the attention matrix $\mathcal{A}\in \mathbb{R}^{(N+1)\times (N+1)}$ by selecting the rows that correspond to the sampled $K' + 1$ tokens. We denote the refined attention matrix by $\mathcal{A}^s\in \mathbb{R}^{(K'+1)\times (N+1)}$. To obtain the output tokens $\mathcal{O}\in \mathbb{R}^{(K'+1)\times d}$, we thus replace the attention matrix $\mathcal{A}$ by the refined one $\mathcal{A}^s$ in \eqref{eq:AttentionMechanism} such that:
\begin{equation}
    \label{eq:RefinedAttentionMechnism}
    \mathcal{O} = \mathcal{A}^{s}\mathcal{V}.
\end{equation}
These output tokens are then taken as input for the next stage. In our experimental evaluation, we demonstrate the efficiency of the proposed adaptive token sampler, which can be added to any vision transformer.     
      
\section{Experiments}

In this section, we analyze the performance of our ATS module by adding it to different backbone models and evaluating them on ImageNet \cite{deng2009imagenet}, Kinetics-400 \cite{kinetics-400}, and Kinetics-600 \cite{kinetics-600}, which are large-scale image and video classification datasets, respectively. In addition, we perform several ablation studies to better analyze our method.
For the image classification task, we evaluate our proposed method on the ImageNet \cite{deng2009imagenet} dataset with 1.3M images and 1K classes. For the action classification task, we evaluate our approach on the Kinetics-400 \cite{kinetics-400} and Kinetics-600 \cite{kinetics-600} datasets with 400 and 600 human action classes, respectively. We use the standard training/testing splits and protocols provided by the ImageNet and Kinetics datasets. If not otherwise stated, the number of output tokens of the ATS module are limited by the number of its input tokens. For example, we set $K=197$ in case of DeiT-S \cite{deit}. For the image classification task, we follow the fine-tuning setup of \cite{dynamicvit} if not mentioned otherwise. The fine-tuned models are initialized by their backbones' pre-trained weights and trained for 30 epochs using PyTorch AdamW optimizer (lr= $\expnumber{5}{-4}$, batch size = $8\times96$). We use the cosine scheduler for training the networks. For more implementation details and also information regarding action classification models, please refer to the supplementary materials.

\subsection{Ablation Experiments}
First, we analyze  different  setups  for  our  ATS  module. Then, we investigate the efficiency and effects of our ATS module when incorporated in different models. If not otherwise stated, we use the pre-trained DeiT-S \cite{deit} model as the backbone and we do not fine-tune the model after adding the ATS module. We integrate the ATS module into stage 3 of the DeiT-S \cite{deit} model. We report the results on the ImageNet-1K validation set in all of our ablation studies.

\begin{figure}[!t]
\begin{center}
   \begin{floatrow}
   \ffigbox[\FBwidth]{\includegraphics[width=0.45\textwidth]{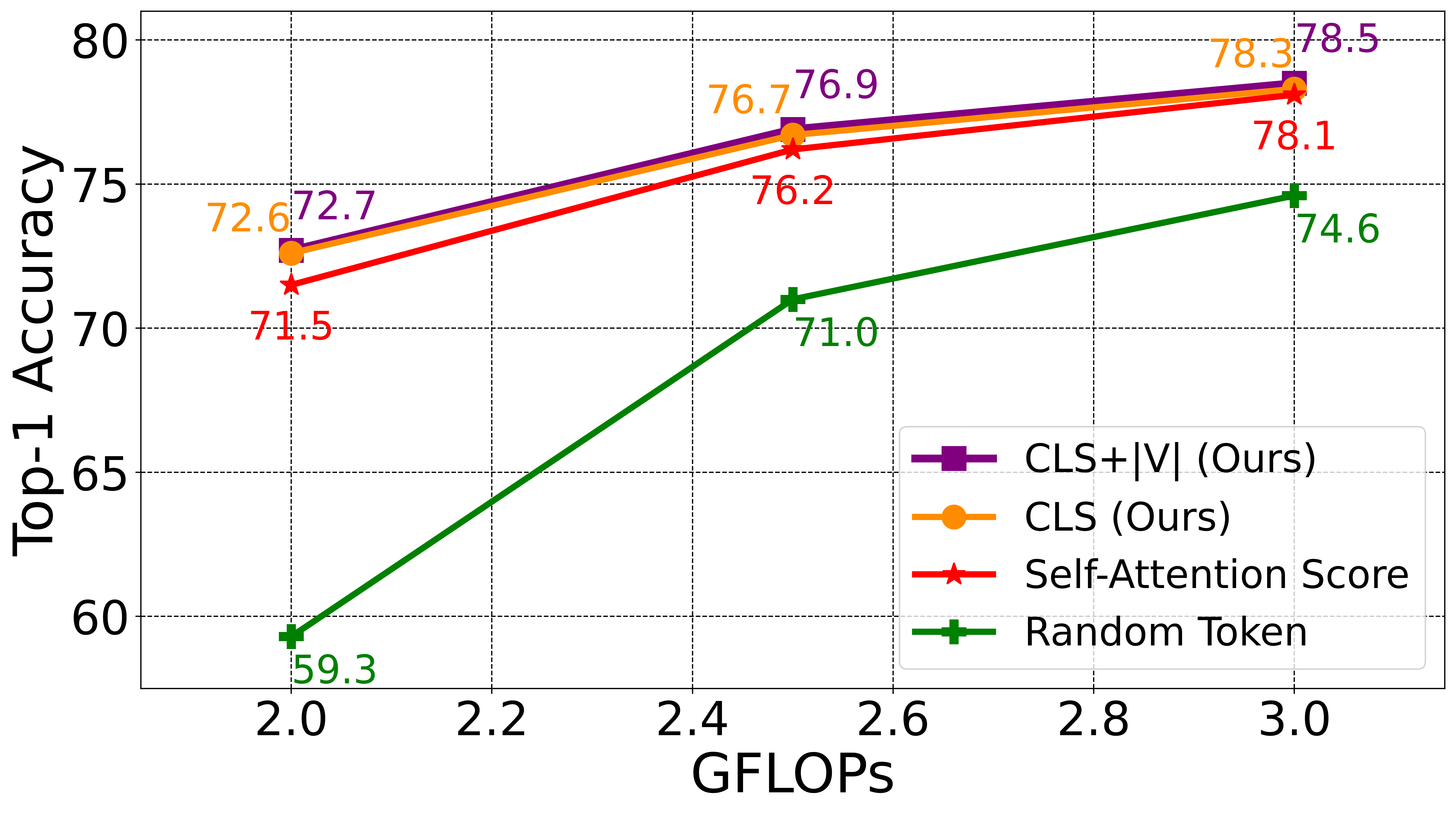}}
   {\caption{Impact of different score assignment methods. To achieve different GFLOPs levels, we bound the value of \(K\) from above such that the average GFLOPs of our adaptive models over the ImageNet validation set reaches the desired level. For more details, please refer to the supplementary material.}
   \label{fig:score_graph}
   }
   
   \hspace{2px}
   
   \ffigbox[\FBwidth]{\includegraphics[width=0.45\textwidth]{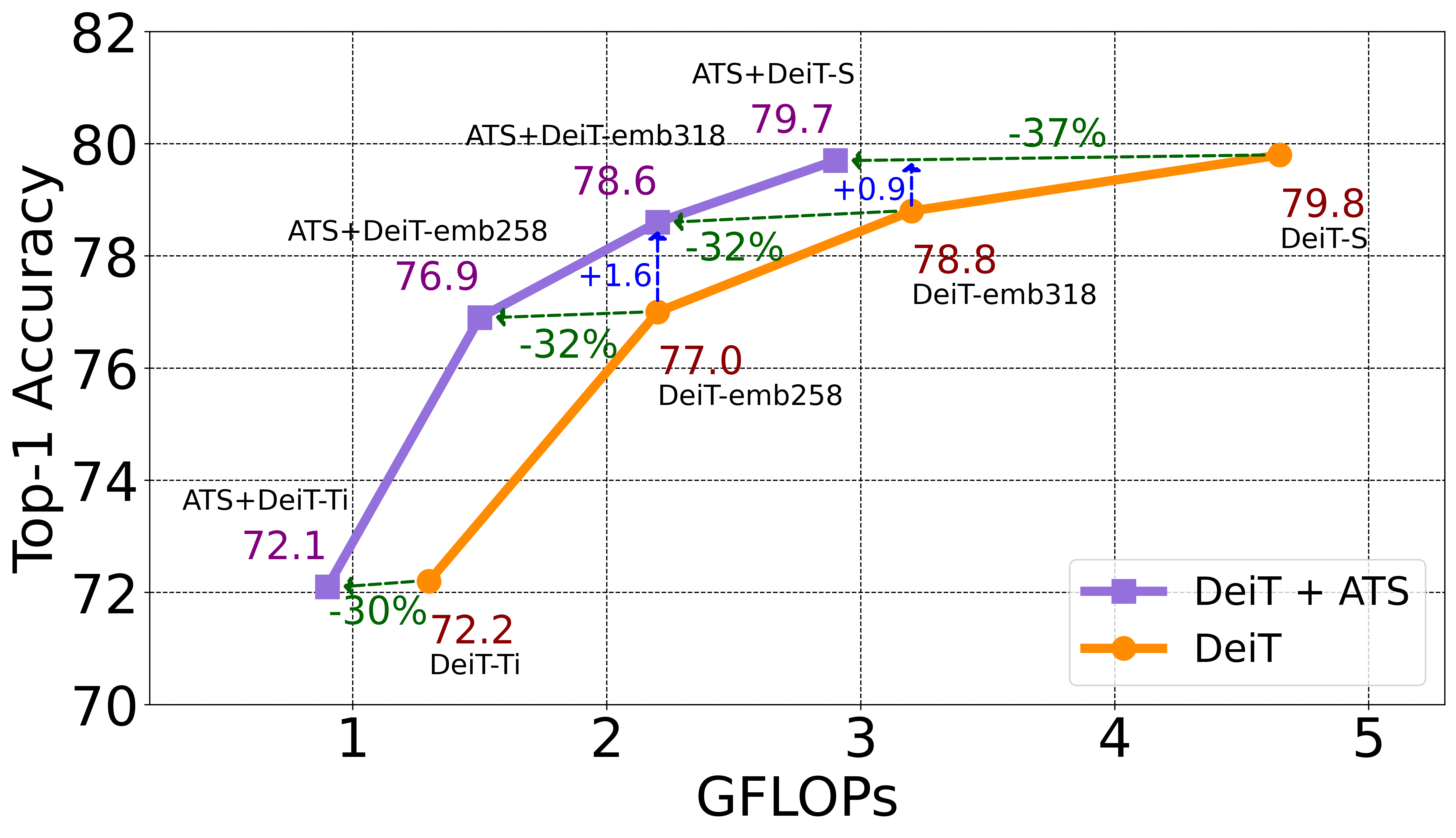}}
   {\caption{\small Performance comparison on the ImageNet validation set. Our proposed adaptive token sampling method achieves a state-of-the-art trade-off between accuracy and GFLOPs. We can reduce the GFLOPs of DeiT-S by $37\%$ while almost maintaining the accuracy.}
   \label{fig:model_scaling}}
   \end{floatrow}
\end{center}
\end{figure}

\noindent \textbf{Significance Scores}~
As mentioned in Sec.~\ref{sec:token_score_assignment}, we use the attention weights of the classification token as significance scores for selecting our candidate tokens. In this experiment, we evaluate different approaches for calculating significance scores. Instead of directly using the attention weights of the classification token, we sum over the attention weights of all tokens (rows of the attention matrix) to find tokens with highest significance over other tokens. We show the results of this method in Fig.~\ref{fig:score_graph} labeled as Self-Attention score. As it can be seen, using the attention weights of the classification token performs better specially in lower FLOPs regimes. The results show that the attention weights of the classification token are a much stronger signal for selecting the candidate tokens. The reason for this is that the classification token will later be used to predict the class probabilities in the final stage of the model. Thus, its corresponding attention weights show which tokens have more impact on the output classification token. Whereas summing over all attention weights only shows us the tokens with highest attention from all other tokens, which may not necessarily be useful for the classification token. 
To better investigate this observation, we also randomly select another token rather than the classification token and use its attention weights for the score assignment. As shown, this approach performs much worse than the other ones both in high and low FLOPs regimes.
We also investigate the impact of using the $L_2$ norm of the values in \eqref{eq:score}. As it can be seen in Fig.~\ref{fig:score_graph}, it improves the results by about $0.2\%$.

\begin{figure*}[ht!]
     \centering
     \subfloat[left][Sampling Methods]
         {\includegraphics[width=0.29\textwidth]{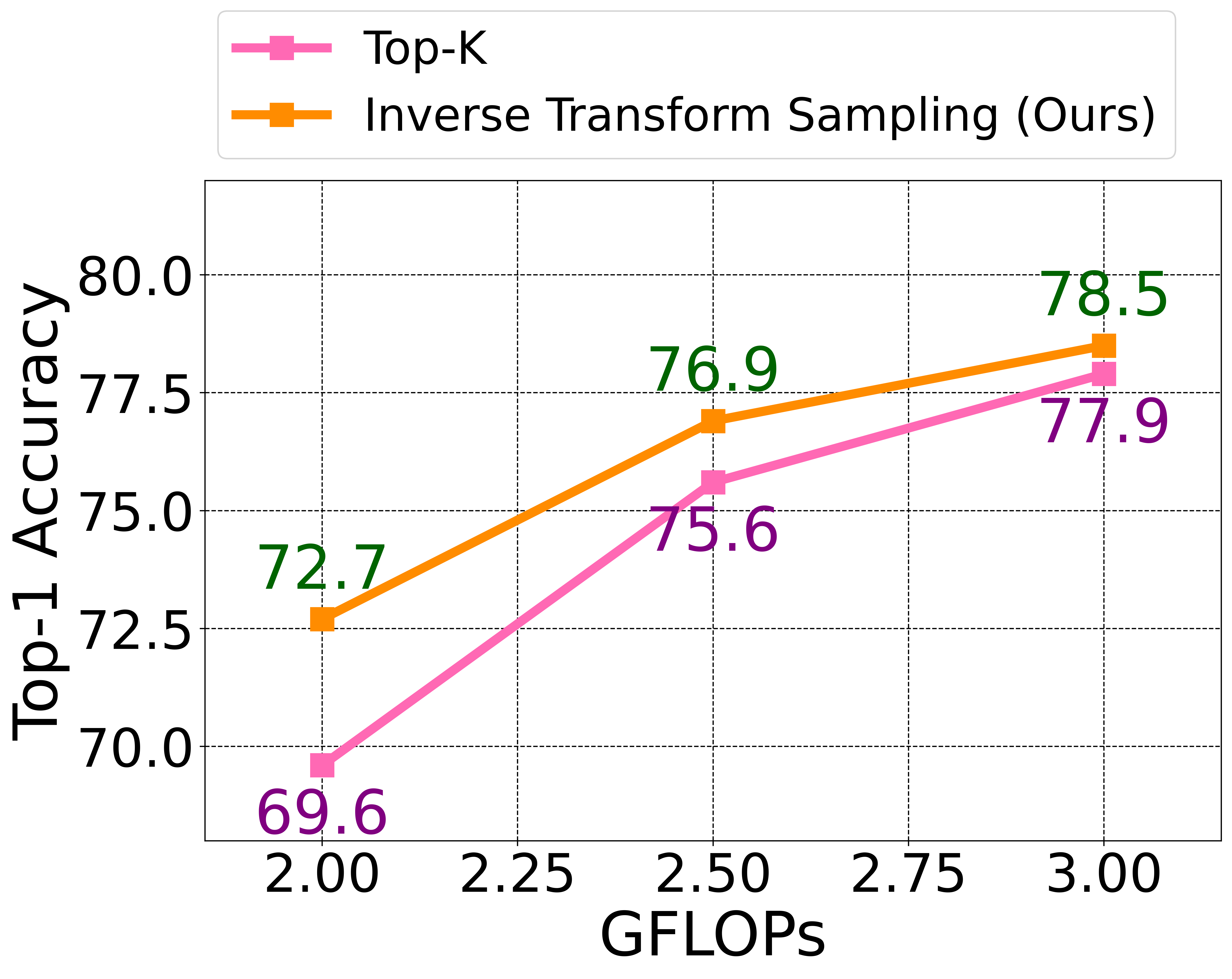}\label{fig:token_sampling}}
     \qquad
     \subfloat[mid][Fine-tuning]
         {\includegraphics[width=0.29\textwidth]{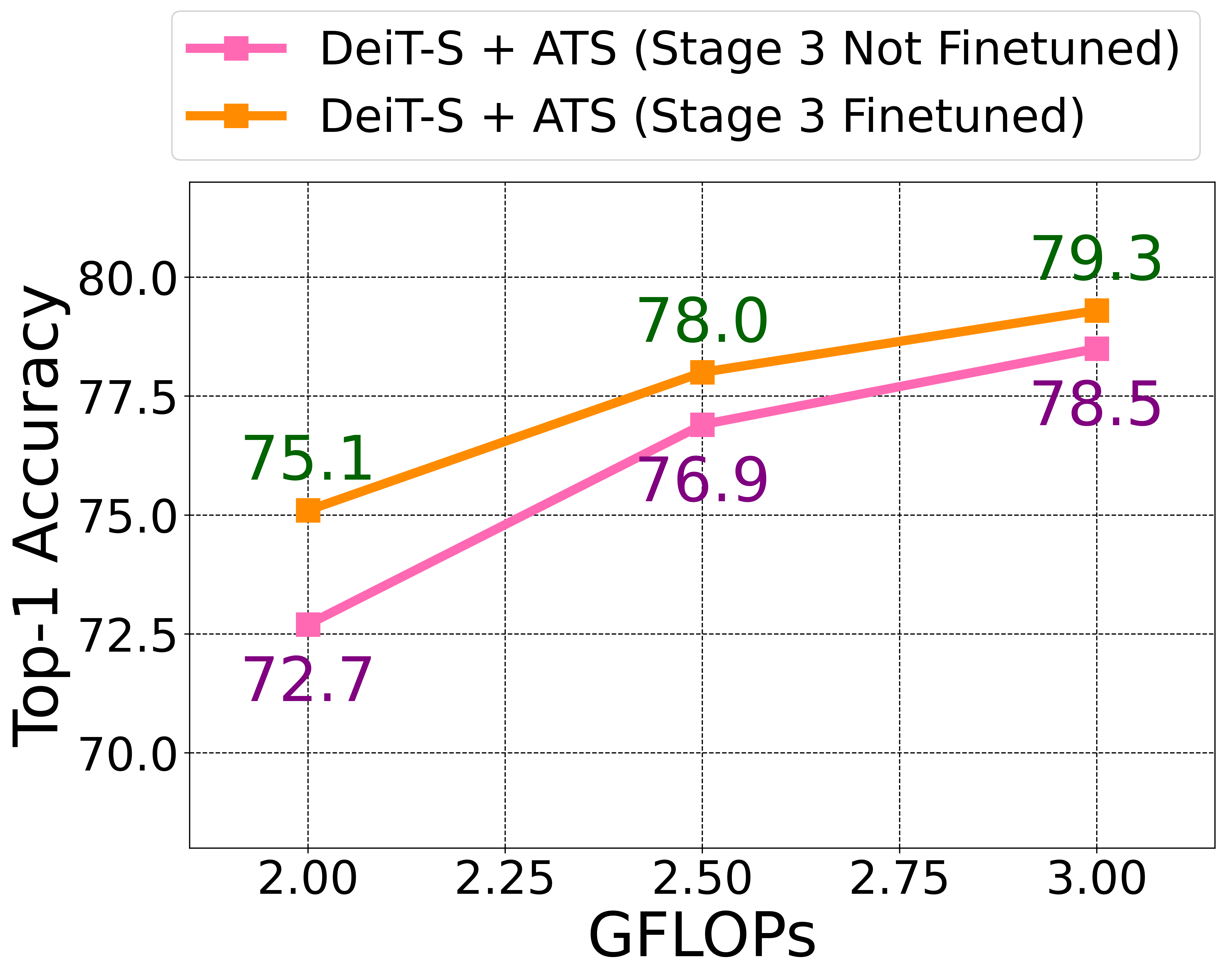}\label{fig:finetuning}}
     \qquad
     \subfloat[right][Multi vs. Single Stage]
         {\includegraphics[width=0.29\textwidth]{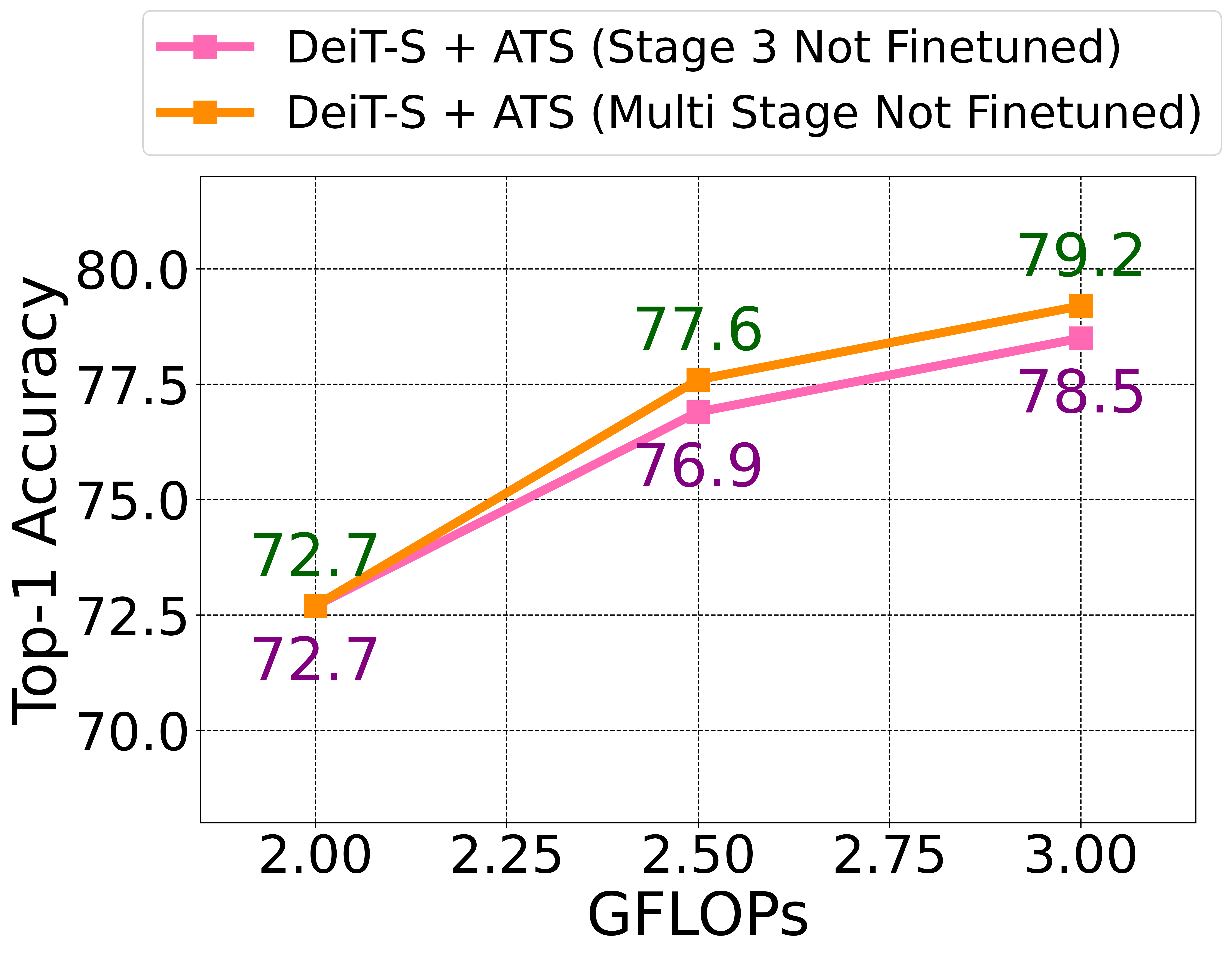}\label{fig:multi_vs_single}}
    \caption{For the model with Top-K selection (fixed-rate sampling) (\ref{fig:token_sampling}), we set $K$ such that the model operates at a desired GFLOPs level. In all three plots, we control the GFLOPs level of our adaptive models as in Fig.~\ref{fig:score_graph}. We use DeiT-S\cite{deit} for these experiments. For more details, please refer to the supplementary material. }
    \label{fig:ablations}
\end{figure*}

\noindent \textbf{Candidate Tokens Selection}~
As mentioned in Sec.~\ref{sec:score_based_token_sampling}, we employ the inverse transform sampling approach to softly downsample the input tokens. To better investigate this approach, we also evaluate the model's performance when picking the top K tokens with highest significance scores $\mathcal{S}$. As it can be seen in Fig.~\ref{fig:token_sampling}, our inverse transform sampling approach outperforms the Top-K selection both in high and low GFLOPs regimes. As discussed earlier, our inverse transform sampling approach based on the CDF of the scores does not hardly discard all tokens with lower significance scores and hence provides a more diverse set of tokens for the following layers. Since earlier transformer blocks are more prone to predict noisier attention weights for the classification token, such a diversified set of tokens can better contribute to the output classification token of the final transformer block. Moreover, the Top-K selection method will result in a fixed token selection rate at every stage that limits the performance of the backbone model. This is shown by the examples in Fig.~\ref{fig:token_removal}. For a cluttered image (bottom), inverse transform sampling keeps a higher number of tokens across all transformer blocks compared to the Top-K selection and hence preserves the accuracy. On the other hand, for a less detailed image (top), inverse transform sampling will retain less tokens, which results in less computation cost.

\noindent \textbf{Model Scaling}~
Another common approach for changing the GFLOPs/accuracy trade-off of networks is to change the channel dimension. To demonstrate the efficiency of our adaptive token sampling method, we thus vary the dimensionality. To this end, we first train several DeiT models with different embedding dimensions. Then, we integrate our ATS module into the stages 3 to 11 of these DeiT backbones and fine-tune the networks. In Fig.~\ref{fig:model_scaling}, we can observe that our approach can reduce GFLOPs by 37\% while maintaining the DeiT-S backbone's accuracy. We can also observe that the GFLOPs reduction rate gets higher as we increase the embedding dimensions from 192 (DeiT-Ti) to 384 (DeiT-S). The results show that our ATS module can reduce the computation cost of the models with larger embedding dimensions to their variants with smaller embedding dimensions.

\noindent \textbf{Visualizations}~
To better understand the way our ATS module operates, we visualize our token sampling procedure (Inverse Transform Sampling) in Fig.~\ref{fig:token_removal}. We have incorporated our ATS module in the stages 3 to 11 of the DeiT-S network. The tokens that are discarded at each stage are represented as a mask over the input image. We observe that our DeiT-S+ATS model has gradually removed irrelevant tokens and sampled those tokens which are more significant to the model's prediction. In both examples, our method identified the tokens that are related to the target objects as the most informative tokens.  

\begin{figure}[t]
\begin{center}
   \begin{floatrow}
   \ffigbox[\FBwidth]{\includegraphics[width=\linewidth]{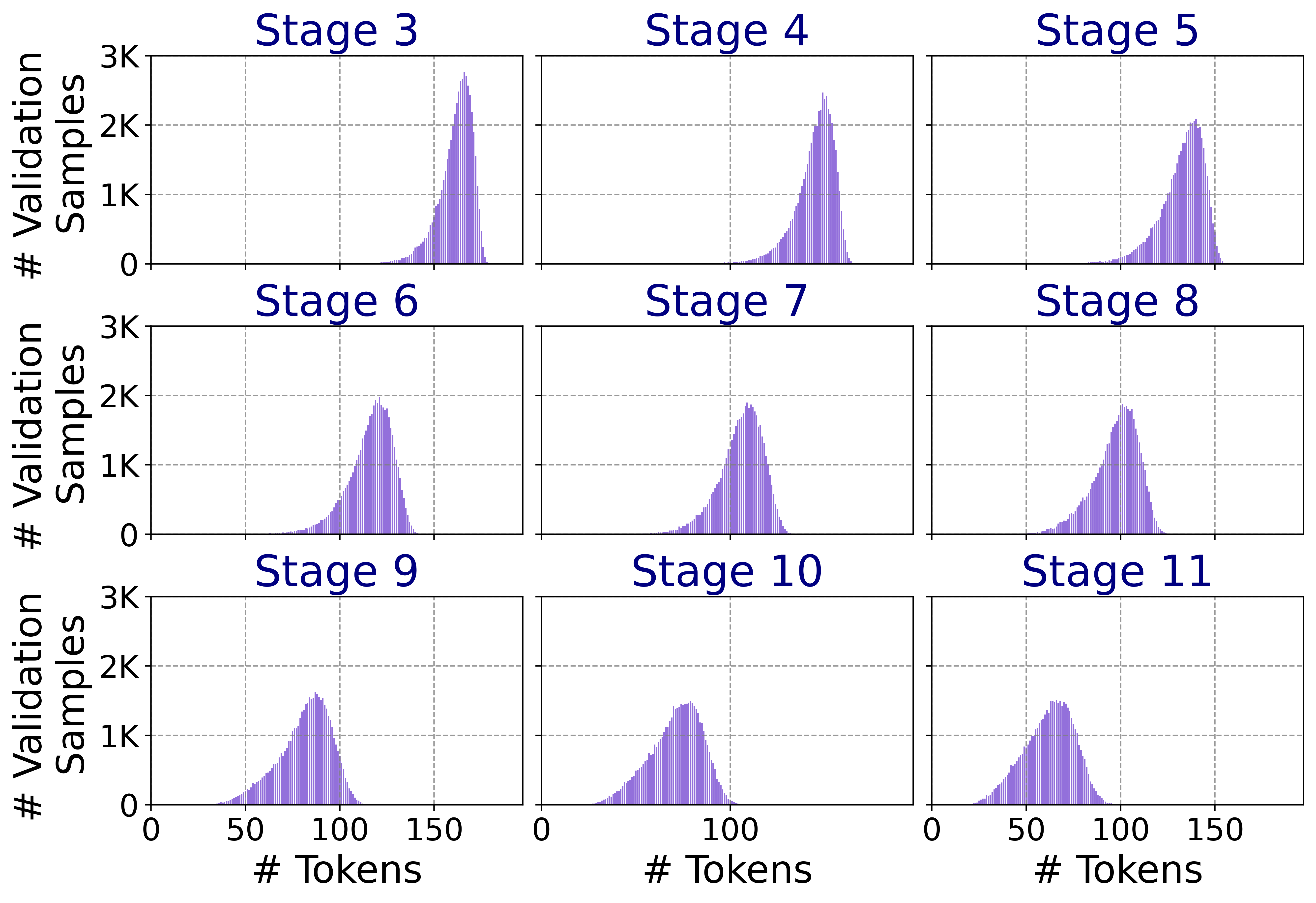}}
   {\caption{Histogram of the number of sampled tokens at each ATS stage of our multi-stage DeiT-S+ATS model on the ImageNet validation set. The y-axis corresponds to the number of images and the x-axis to the number of sampled tokens.}\label{fig:dynamicity}}
   \ffigbox[0.9\FBwidth]{\includegraphics[width=0.49\linewidth]{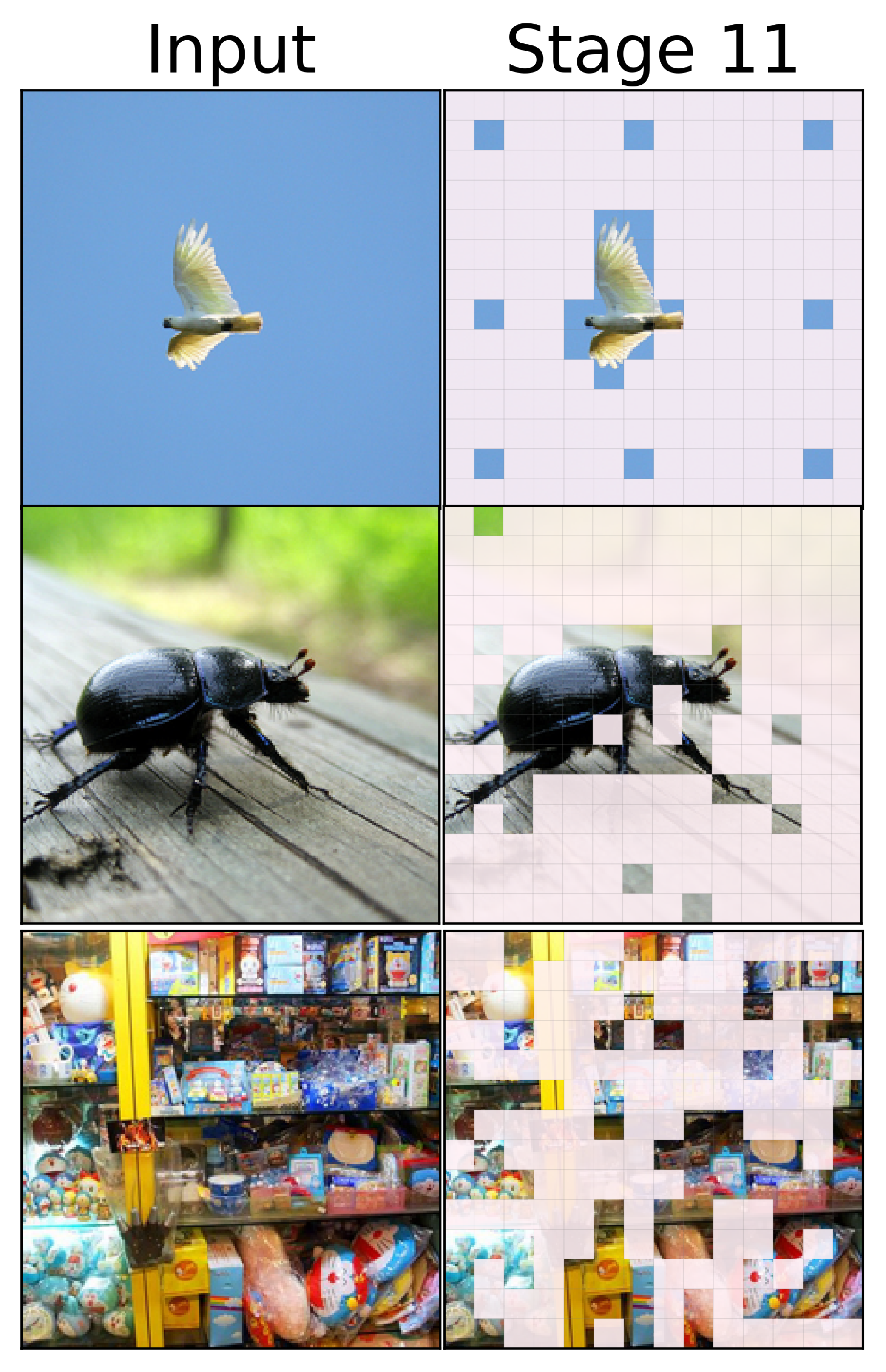}%
                      \includegraphics[width=0.49\linewidth]{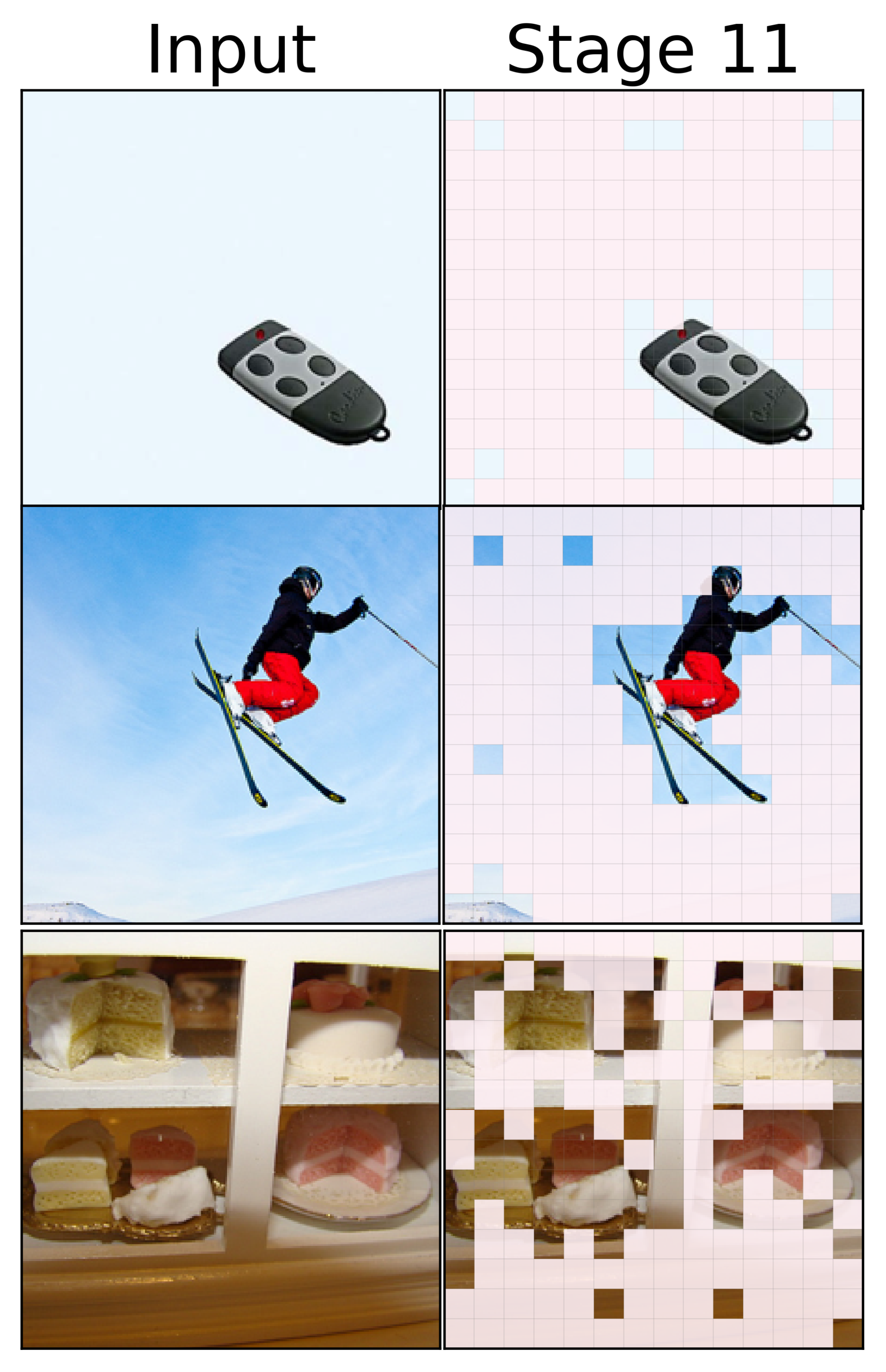}}
   {\caption{ ATS samples less tokens for images with fewer details (top), and a higher number of tokens for more detailed images (bottom). We show the token downsampling results after all ATS stages. For this experiment, we use a multi-stage Deit-S+ATS model.} 
   \label{fig:dynamicity_visualization}}
   \end{floatrow}
\end{center}
\end{figure}

\noindent \textbf{Adaptive Sampling}~
In this experiment, we investigate the adaptivity of our token sampling approach. We evaluate our multi-stage DeiT-S+ATS model on the ImageNet validation set. In Fig.~\ref{fig:dynamicity}, we visualize histograms of the number of sampled tokens at each ATS stage. We can observe that the number of selected tokens varies at all stages and for all images. We also qualitatively analyze this nice property of our ATS module in Figs.~\ref{fig:token_removal} and \ref{fig:dynamicity_visualization}. We can observe that our ATS module selects a higher number of tokens when it deals with detailed and complex images while it selects a lower number of tokens for less detailed images.

\noindent \textbf{Fine-tuning}~
To explore the influence of fine-tuning on the performance of our approach, we fine-tune a DeiT-S+ATS model on the ImageNet training set. We compare the model with and without fine-tuning. As shown in Fig.~\ref{fig:finetuning}, fine-tuning can improve the accuracy of the model. In this experiment, we fine-tune the model with $K=197$ but test it with different $K$ values to reach the desired GFLOPs levels.

\noindent \textbf{ATS Stages}~
In this experiment, we explore the effect of single-stage and multi-stage integration of the ATS block into vision transformer models. In the single-stage model, we integrate our ATS module into the stage 3 of DeiT-S. In the multi-stage model, we integrate our ATS module into the stages 3 to 11 of DeiT-S. As it can be seen in Fig.~\ref{fig:multi_vs_single}, the multi-stage DeiT-S+ATS performs better than the single-stage DeiT-S+ATS. This is due to the fact that a multi-stage DeiT-S+ATS model can gradually decrease the GFLOPs by discarding fewer tokens in earlier stages, while a single-stage DeiT-S+ATS model has to discard more tokens in earlier stages to reach the same GFLOPs level.

\subsection{Comparison with state-of-the-art}
We compare the performances of our adaptive models, which are equipped with the ATS module, with state-of-the-art vision transformers for image and video classification on the ImageNet-1K \cite{deng2009imagenet} and Kinetics \cite{kinetics-400, kinetics-600} datasets, respectively. Tables~\ref{tab:imagenet}-\ref{tab:k600} show the results of this comparison. For the image classification task, we incorporate our ATS module into the stages 3 to 11 of the DeiT-S \cite{deit} model. We also integrate our ATS module into the \(1^{st}\) to \(9^{th}\) blocks of the \(3^{rd}\) stage of CvT-13 \cite{cvt} and CvT-21 \cite{cvt}, and into stages 1-9 of the transformer module of PS-ViT~\cite{psvit}. We fine-tune the models on the ImageNet-1K training set.  We also evaluate our ATS module for action recognition. To this end, we add our module to the XViT~\cite{xvit} and TimeSformer~\cite{bertasius2021space} video vision transformers. For more details, please refer to the supplementary materials.

\noindent\textbf{Image Classification}
As it can be seen in Table~\ref{tab:imagenet}, our ATS module decreases the GFLOPs of all vision transformer models without adding any extra parameters to the backbone models. For the DeiT-S+ATS model, we observe a $37\%$ GFLOPs reduction with only $0.1\%$ reduction of the top-1 accuracy. For the CvT+ATS models, we can also observe a GFLOPs reduction of about $30\%$ with $0.1-0.2\%$ reduction of the top-1 accuracy. More details on the efficiency of our ATS module can be found in the supplementary materials (\eg throughput). 
Comparing ATS to DynamicViT \cite{dynamicvit} and HVT \cite{HVT}, which add additional parameters to the model, our approach achieves a better trade-off between accuracy and GFLOPs. Our method also outperforms the EViT-DeiT-S~\cite{evit} model trained for 30 epochs without adding any extra trainable parameters to the model. We note that the EViT-DeiT-S model can improve its top-1 accuracy by around 0.3\% when it is trained for much more training epochs (\eg 100 epochs). For a fair comparison, we have considered the 30 epochs training setup used by Dynamic-ViT~\cite{dynamicvit}. We have also added our ATS module to the PS-ViT network~\cite{psvit}. As it can be seen in Table~\ref{tab:imagenet}, although PS-ViT has drastically lower GFLOPs compared to its counterparts, its GFLOPs can be further decreased by incorporating ATS in it.

\noindent \textbf{Action Recognition}
As it can be seen in Tables~\ref{tab:k400} and \ref{tab:k600}, our {ATS} module drastically decreases the GFLOPs of all video vision transformers without adding any extra parameters to the backbone models. For the XViT+ATS model, we observe a $39\%$ GFLOPs reduction with only $0.2\%$ reduction of the top-1 accuracy on Kinetics-400 and a $38.7\%$ GFLOPs reduction with only $0.1\%$ drop of the top-1 accuracy on Kinetics-600. We observe  that XViT+ATS achieves a similar accuracy as TokenLearner~\cite{tokenlearner} on Kinetics-600 while requiring $17.6\times$ less GFLOPs. For TimeSformer-L+ATS, we can observe $50.8\%$ GFLOPs reduction with only $0.2\%$ drop of the top-1 accuracy on Kinetics-400. These results demonstrate the generality of our approach that can be applied to both image and video representations.

\begin{table}[tb]
\RawFloats
  \centering
  \begin{minipage}{0.49\linewidth}
    \caption{Comparison of the multi-stage ATS models with state-of-the-art image classification models with comparable GFLOPs on the ImageNet validation set. We equip DeiT-S \cite{deit}, PS-ViT~\cite{psvit}, and variants of CvT \cite{cvt} with our ATS module and fine-tune them on the ImageNet training set.} \vspace{10px}
    \resizebox{\textwidth}{!}{
    \begin{tabular}{lcccc}\toprule
    Model & Params (M) & GFLOPs & Resolution   & Top-1 \\\midrule
    ViT-Base/16~\cite{vit} & 86.6  & 17.6  & 224   & 77.9  \\\midrule
    HVT-S-1~\cite{HVT} & 22.09 & 2.4 & 224 & 78.0 \\ 
    IA-RED$^2$~\cite{IARED} & - & 2.9 & 224 & 78.6 \\
    DynamicViT-DeiT-S (30 Epochs)~\cite{dynamicvit} & 22.77  & 2.9   & 224   & 79.3  \\
    EViT-DeiT-S (30 epochs)~\cite{evit} & 22.1 & 3.0 & 224 & 79.5 \\
    \rowcolor{LightYellow}
    DeiT-S+ATS (Ours) & \textbf{22.05}  & \textbf{2.9}   & 224   & 79.7  \\
    DeiT-S~\cite{deit} & 22.05  & 4.6   & 224   & 79.8  \\\midrule
    
    PVT-Small~\cite{wang2021pvt} & 24.5  & 3.8   & 224   & 79.8  \\
    CoaT Mini~\cite{xu2021coat} & 10.0  & 6.8   & 224   & 80.8  \\
    CrossViT-S~\cite{chen2021crossvit} & 26.7  & 5.6   & 224   & 81.0  \\
    PVT-Medium~\cite{wang2021pvt} & 44.2  & 6.7   & 224   & 81.2  \\
    Swin-T~\cite{swin} & 29.0  & 4.5   & 766   & 81.3  \\
    T2T-ViT-14~\cite{yuan2021t2t} & 22.0  & 5.2   & 224   & 81.5  \\
    CPVT-Small-GAP~\cite{chu2021cpvt} & 23.0  & 4.6   & 817   & 81.5  \\\midrule
    
    CvT-13~\cite{cvt} & 20.0  & 4.5  & 224   & 81.6  \\
    \rowcolor{LightYellow}
    CvT-13+ATS (Ours) & 20.0  & \textbf{3.2}  & 224   & 81.4  \\\midrule
    PS-ViT-B/14~\cite{psvit} & 21.3  &  5.4 & 224 & 81.7  \\
    \rowcolor{LightYellow}
    PS-ViT-B/14+ATS (Ours) & 21.3  & \textbf{3.7} & 224 & 81.5  \\\midrule
    RegNetY-8G~\cite{radosavovic2020designing} & 39.0  & 8.0   & 224   & 81.7  \\
    DeiT-Base/16~\cite{deit} & 86.6  & 17.6  & 224   & 81.8  \\
    CoaT-Lite Small~\cite{xu2021coat} & 20.0  & 4.0   & 224   & 81.9  \\
    T2T-ViT-19~\cite{yuan2021t2t} & 39.2  & 8.9   & 224   & 81.9  \\
    
    
    CrossViT-B~\cite{chen2021crossvit} & 104.7  & 21.2  & 224   & 82.2  \\
    T2T-ViT-24~\cite{yuan2021t2t} & 64.1  & 14.1  & 224   & 82.3  \\\midrule
    PS-ViT-B/18~\cite{psvit} & 21.3  & 8.8 & 224 & 82.3  \\
    \rowcolor{LightYellow}
    PS-ViT-B/18+ATS (Ours) & 21.3  & \textbf{5.6} & 224 & 82.2  \\\midrule
    CvT-21~\cite{cvt} & 32.0  & 7.1  & 224   & 82.5  \\
    \rowcolor{LightYellow}
    CvT-21+ATS (Ours) & 32.0  & \textbf{5.1}  & 224   & 82.3  \\\midrule
    
    TNT-B~\cite{han2021transformer} & 66.0  & 14.1  & 224   & 82.8  \\
    RegNetY-16G~\cite{radosavovic2020designing} & 84.0  & 16.0  & 224   & 82.9  \\
    Swin-S~\cite{swin} & 50.0  & 8.7   & 224   & 83.0  \\\midrule
    
    CvT-13$_{384}$~\cite{cvt} & 20.0  & 16.3  & 384   & 83.0  \\
    \rowcolor{LightYellow}
    CvT-13$_{384}$+ATS (Ours) & 20.0  & \textbf{11.7}  & 384   & 82.9  \\\midrule
    
    Swin-B~\cite{swin} & 88.0  & 15.4  & 224   & 83.3  \\
    LV-ViT-S~\cite{jiang2021token} & 26.2 & 6.6 & 224 & 83.3 \\\midrule
    
    
    CvT-21$_{384}$~\cite{cvt} & 32.0  & 24.9  & 384   & 83.3  \\
    \rowcolor{LightYellow}
    CvT-21$_{384}$+ATS (Ours) & 32.0  & \textbf{17.4}  & 384   & 83.1  \\\bottomrule
    \end{tabular}}
    \label{tab:imagenet}
  \end{minipage}%
  \hspace{2px}
  \begin{minipage}{0.45\linewidth}
    \caption{Comparison with state-of-the-art on Kinetics-400.}
    \vspace{10px}
    \resizebox{\textwidth}{!}{
    \begin{tabular}{lcccc}\toprule
    Model & Top-1 & Top-5 & Views & GFLOPs \\ \midrule
    STC~\cite{diba2018spatio} & 68.7 & 88.5 & 112 & -\\
    bLVNet~\cite{fan2019blvnet} & 73.5 & 91.2 & 3$\times$3 & 840 \\
    STM~\cite{lin2019tsm} & 73.7 & 91.6 & - & - \\
    TEA~\cite{li2020tea} & 76.1 & 92.5 & 10$\times$3 & 2,100 \\
    TSM R50~\cite{jiang2019stm} & 74.7 & - & 10$\times$3 & 650 \\
    I3D NL~\cite{wang2018non} & 77.7 & 93.3 & 10$\times$3 & 10,800 \\
    CorrNet-101~\cite{wang2020video} & 79.2 & - & 10$\times$3 & 6,700 \\
    ip-CSN-152~\cite{tran2019video} & 79.2 & 93.8 & 10$\times$3 & 3,270 \\
    HATNet \cite{diba2020large} & 79.3 & - & - & - \\
    SlowFast 16$\times$8 R101+NL~\cite{feichtenhofer2019slowfast} & 79.8 & 93.9 & 10$\times$3 & 7,020 \\
    X3D-XXL~\cite{feichtenhofer2020x3d} & 80.4 & 94.6 & 10$\times$3 & 5,823 \\\midrule
    TimeSformer-L~\cite{bertasius2021space} & 80.7 & 94.7 & 1$\times$3 & 7,140 \\
    \rowcolor{LightYellow}
    TimeSformer-L+ATS (Ours) & 80.5 & 94.6 & 1$\times$3 & \textbf{3,510} \\\midrule
    ViViT-L/16x2~\cite{bertasius2021space} & 80.6 & 94.7 & 4$\times$3 & 17,352 \\
    MViT-B, 64×3 \cite{fan2021multiscale} & 81.2 & 95.1 & 3$\times$3 & 4,095\\\midrule
    X-ViT (16$\times$) \cite{xvit} & 80.2 & 94.7 & 1$\times$3 & 425 \\
    \rowcolor{LightYellow}
    X-ViT+ATS (16$\times$) (Ours) & 80.0 & 94.6 & 1$\times$3 & \textbf{259}\\\midrule
    TokenLearner 16at12 (L/16) \cite{tokenlearner} & 82.1 & - & 4$\times$3 & 4,596
    \\ \bottomrule
    \label{tab:k400}
    \end{tabular}}
    \vspace{5px}
    \caption{Comparison with state-of-the-art on Kinetics-600.}
    \vspace{10px}
    \resizebox{\textwidth}{!}{
    \begin{tabular}{lcccc}\toprule
    Model & Top-1 & Top-5 & Views & GFLOPs \\ \midrule
    AttentionNAS \cite{attentionnas} & 79.8 & 94.4 & - & 1,034 \\
    LGD-3D R101 \cite{qiu2019learning} & 81.5 & 95.6 & 10$\times$3 & - \\
    HATNET \cite{diba2020large} & 81.6 & - & - & - \\
    SlowFast R101+NL \cite{feichtenhofer2019slowfast} & 81.8 & 95.1 & 10$\times$3 & 3,480 \\
    X3D-XL \cite{feichtenhofer2020x3d} & 81.9 & 95.5 & 10$\times$3 & 1,452 \\
    X3D-XL+ATFR \cite{atfr} & 82.1 & 95.6 & 10$\times$3 & 768 \\\midrule
    TimeSformer-HR \cite{bertasius2021space} & 82.4 & 96 & 1$\times$3 & 5,110 \\
    \rowcolor{LightYellow}
    TimeSformer-HR+ATS (Ours) & 82.2 & 96 & 1$\times$3 & \textbf{3,103} \\\midrule
    ViViT-L/16x2 \cite{bertasius2021space} & 82.5 & 95.6 & 4$\times$3 & 17,352 \\
    Swin-B \cite{swin} & 84.0 & 96.5 & 4$\times$3 & 3,384\\
    MViT-B-24, 32×3 \cite{fan2021multiscale} & 84.1 & 96.5 & 1$\times$5 & 7,080 \\
    TokenLearner 16at12(L/16) \cite{tokenlearner} & 84.4 & 96.0 & 4$\times$3 & 9,192 \\\midrule
    X-ViT (16$\times$) \cite{xvit} & 84.5 & 96.3 & 1$\times$3 & 850 \\
    \rowcolor{LightYellow}
    X-ViT+ATS (16$\times$) (Ours) & 84.4 & 96.2 & 1$\times$3 & \textbf{521} 
    \\ \bottomrule
    \label{tab:k600}
    \end{tabular}}
  \end{minipage}
\end{table}%

\section{Conclusion}
Designing computationally efficient vision transformer models for image and video recognition is a challenging task.
In this work, we proposed a novel differentiable parameter-free module called Adaptive Token Sampler (ATS) to increase the efficiency of vision transformers for image and video classification. The new ATS module selects the most informative and distinctive tokens within the stages of a vision transformer model such that as much tokens as needed but not more than necessary are used for each input image or video clip.
By integrating our ATS module into the attention layers of current vision transformers, which use a static number of tokens, we can convert them into much more efficient vision transformers with an adaptive number of tokens. We showed that our ATS module can be added to off-the-shelf pre-trained vision transformers as a plug and play module, thus reducing their GFLOPs without any additional training, but it is also possible to train a vision transformer equipped with the ATS module thanks to its differentiable design.
We evaluated our approach on the ImageNet-1K image recognition dataset and incorporated our ATS module into three different state-of-the-art vision transformers. We also demonstrated the generality of our approach by incorporating it into different state-of-the-art video vision transformers and evaluating them on the Kinetics-400 and Kinetics-600 datasets. The results show that the ATS module decreases the computation cost (GFLOPs) between $27\%$ and $50.8\%$ with a negligible accuracy drop. Although our experiments are focused on image and video vision transformers, we believe that our approach can also work in other domains such as audio.

\textbf{Acknowledgments:} Farnoush Rezaei Jafari acknowledges support by the Federal Ministry of Education and Research (BMBF) for the Berlin Institute for the Foundations of Learning and Data (BIFOLD) (01IS18037A). Juergen Gall has been supported by the Deutsche Forschungsgemeinschaft (DFG, German Research Foundation) under Germany's Excellence Strategy - EXC 2070 - 390732324, GA1927/4-2 (FOR 2535 Anticipating Human Behavior), and the ERC Consolidator Grant FORHUE (101044724).

\newcount\cvprrulercount
\appendix
\section*{Appendix}

\setcounter{table}{0}
\setcounter{figure}{0}
\renewcommand{\thetable}{A.\arabic{table}}
\renewcommand{\thefigure}{A.\arabic{figure}}

\section{Runtime}

\noindent\textbf{Throughput:} While ATS is a super-light module, there is still a small cost associated with I/O operations. For a DeiT-S network with a single ATS stage, the sampling overhead is about $1.5\%$ of the overall computation which is negligible compared to the large savings due to the dropped tokens. To further elaborate on this, we have reported the throughput (images/s) of the DeiT-S model with/without our ATS module in Table~\ref{tab:throughput}. As it can be seen, the speed-up of our module is aligned with its GFLOPs reduction. \\

\noindent\textbf{Batch Processing:} While for most applications the inference is performed for a single image or video, ATS can also be used for inference with a mini-batch. To this end, we rearrange the tokens of each image so that the sampled tokens are in the lower indices. Then, we remove the last tokens completely to reduce the computation. This way, we only process $m$ tokens, where $m=\max_i(K_i'+1)$ over all images $i$ of the mini-batch. In the worst case scenario (\eg a very large minibatch), we will keep all $K+1$ first tokens after rearrangement. This will still reduce the computation by a factor of $\frac{N+1}{K+1}$. For example, using a mini-batch of size 512 on the ImageNet validation set, $m$ is $129$ in Stage $7$ of the DeiT-S+ATS model, which is smaller than the total number of tokens ($197$). Therefore, we discard at least $68$ tokens in stage $7$ even in a mini-batch setting. Moreover, for the fully connected layers in a transformer block, which requires most of the computation \cite{token-pooling}, we can flatten the mini-batch dimension and forward only non-zero tokens of the whole mini-batch in parallel through the fully connected layers.

\section{The Effect of K}
\begin{figure}[t]
\RawFloats
\begin{minipage}{1.0\textwidth}
  \centering
  \begin{minipage}[b]{0.45\textwidth}
    \captionof{table}
    {\textbf{Runtime comparison:} We run the models on a single RTX6000 GPU (CUDA 11.0, PyTorch 1.8, image size: 224$\times$224)
    . We average the value of throughput over 20 runs. We add ATS to multiple stages of the DeiT-S model and fine-tune the network on the ImageNet dataset.}
    \vspace{+0.1in}
    \scalebox{0.7}{
    \begin{tabular}{lccccc}\toprule
    Model & Params (M) & GFLOPs & Throughput & Top-1 \\\midrule
    Deit-S~\cite{deit} & 22.05  & 4.6  & 1010  & 79.8 \\\midrule
    Deit-S+ATS & 22.05  & \textbf{2.9}  & \textbf{1403} & 79.7  \\\midrule
    \end{tabular}
    }
    \label{tab:throughput}
    \end{minipage}
    \hfill
    \begin{minipage}[b]{0.49\textwidth}
    \centering
     \includegraphics[width=0.8\linewidth]{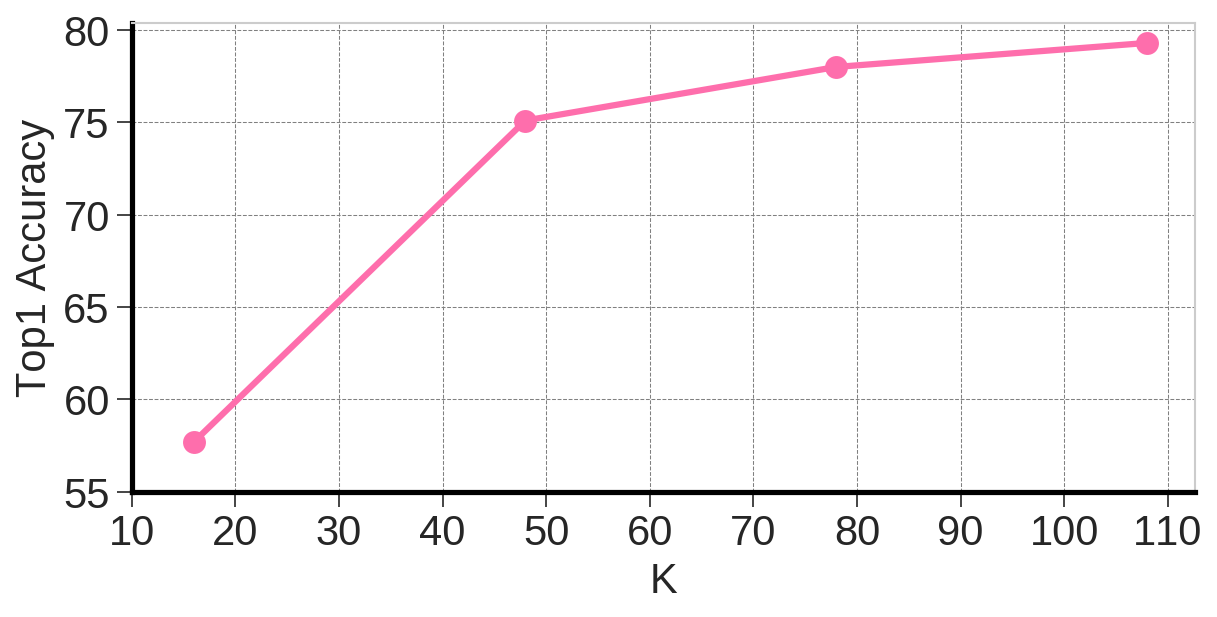}
     \vspace{-0.1in}
    \captionof{figure}{\textbf{Effect of K:} We varied the value of K in the ATS module to study the effect of K on the top-1 accuracy. K=48 corresponds to 2 GFLOPs. The backbone model is DeiT-S pre-trained on ImageNet-1K.}
   \label{fig:ablate_k}
  \end{minipage}
\end{minipage}
\end{figure}

In Fig.~5 of the main paper, we varied the value of $K$ to achieve different GFLOPs levels (Top-1 Accuracy vs.\ GFLOPs). In Fig.~\ref{fig:ablate_k}, we study the effect of varying K in the ATS module of the single-stage DeiTS+ATS model with fine-tuning. Interestingly, even sampling only 48 tokens (2 GFLOPs) achieves 75\% accuracy.


\section{ATS Integration Without Further Training}
One of the most important aspects of our approach is that it can be added to any pre-trained off-the-shelf vision transformer. For example, our not fine-tuned multi-stage DeiT-S+ATS model (Fig.~5(c) in the paper) has only a 0.6\% (Table 1 in the paper) top-1 accuracy  drop while it has improved the efficiency by about $1.6$ GFLOPs without any further training of the backbone model. We also observe the same performance on video data. As reported in Table~\ref{tab:xvit+ats_notfinetuned}, our not fine-tuned XViT+ATS model has only a 1.1\% top-1 accuracy drop while it has improved the efficiency by about $329$ GFLOPs without any further training of the backbone model. This capability of our ATS module roots back in its adaptive inverse transform sampling strategy. Our ATS module samples informative tokens based on their contributions to the classification token. Uninformative tokens that only slightly contribute to the final prediction receive lower attention weights for the classification token. Therefore, the output classification token will be only marginally affected by removing such redundant tokens. On the other hand, the redundant tokens are less similar to the informative tokens and receive lower attention weights for those tokens in the attention matrix. Consequently, they do not contribute much to the value of informative tokens and eliminating them does not change the way informative tokens are contributing to the output classification token.

\begin{minipage}[b]{1.0\textwidth}
\centering
\captionof{table}{Our ATS module is added to XViT~\cite{xvit} pre-trained on Kinetics-600. 
}
\vspace{+0.1in}
\scalebox{1.0}{
\begin{tabular}{lcc}
    \toprule 
    Model & Top-1 & GFLOPs  \\

    \midrule
    XViT+ATS Not-Finetuned(16$\times$) & 83.4 & \textbf{521} \\
    XViT+ATS Finetuned(16$\times$) & 84.4 & \textbf{521} \\ 
    XViT(16$\times$) & \textbf{84.5} & 850 \\ 
    \bottomrule
    \end{tabular}
}
\label{tab:xvit+ats_notfinetuned}
\end{minipage}

\section{Attention Map Visualization}
\begin{figure}
    \centering
    \includegraphics[width=0.9\linewidth]{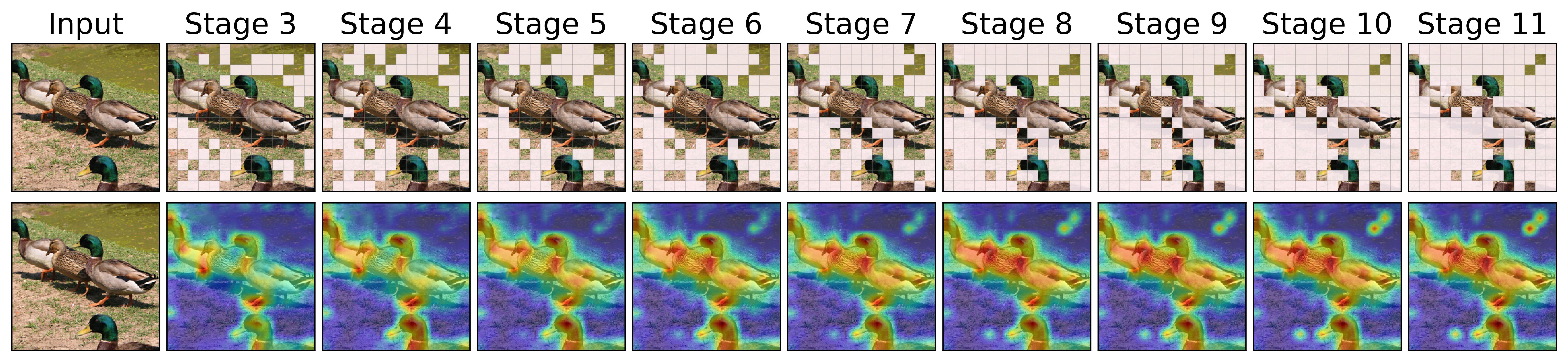}
    \caption{Visualization of the sampled tokens and attention maps of a not fine-tuned multi-stage DeiT-S+ATS. 
    }
    \label{fig:attention_visualization}
\end{figure}

As shown in Fig. \ref{fig:attention_visualization}, the attention maps become more focused on the birds and less on the background at the later stages, which is aligned with our observations on the sampled tokens at each stage.

\section{Implementation Details}
In our experiments for image classification, we use the ImageNet \cite{deng2009imagenet} dataset with 1.28M training images and 1K classes. We evaluate our adaptive models, which are equipped with the ATS module, on 50K validation images of this dataset.  In our experiments for action recognition, we use the Kinetics-400 \cite{kinetics-400} and Kinetics-600 \cite{kinetics-600} datasets containing short clips (typically 10 seconds long) sampled from YouTube. Kinetics-400 and Kinetics-600 consist of 400 and 600 classes, respectively. The versions of Kinetics-400 and Kinetics-600 used in this paper consist of approximately 261k and 457k clips, respectively. Note that these numbers are lower than the original datasets due to the removal of certain videos from YouTube. Our networks for image classification are trained on 8 NVIDIA Quadro RTX 6000 GPUs and for action recognition on 8 NVIDIA A100 GPUs.

\subsection{DeiT + ATS}
\noindent\textbf{Training}
To fine-tune our adaptive models, we follow the DynamicViT \cite{dynamicvit} training settings. We use the DeiT model's pre-trained weights to initialize the backbones of our adaptive network and train it for 30 epochs using AdamW optimizer. The learning rate and batch size are set to 5e-4 and $8\times96$, respectively. We use the cosine scheduler to train the networks. For both multi and single stage models, we set $K=197$ during training.
\vspace{2px}

\noindent\textbf{Evaluation}
We use the same setup as \cite{deit} for evaluating our adaptive models. To evaluate the performance of our multi-stage DeiT-S+ATS model with different average GFLOPs levels of $3$, $2.5$, and $2$, we set $K_n=\max(\round{\rho\times\#InputTokens_n}, 8)$ in which $\rho$ is set to $1$, $0.87$, $0.72$, respectively, and $n$ is the stage index. For the single-stage model, we set $K=108$, $78$, $48$ to evaluate the model with different average GFLOPs levels of $3$, $2.5$, and $2$.

\subsection{CvT + ATS}
We integrate our ATS module into the \(1^{st}\) to \(9^{th}\) blocks of the \(3^{rd}\) stage of the CvT-13 \cite{cvt} and CvT-21 \cite{cvt} networks. For both CvT models, we do not use any convolutional projection layers in the transformer blocks of stage 3.
\vspace{2px}

\noindent\textbf{Training}
To train our adaptive models, we follow most of the CvT \cite{cvt} network's training settings. We use the CvT model's pre-trained weights to initialize the backbones of our adaptive networks and train them for 30 epochs using AdamW optimizer. The learning rate and batch size are set to 1.5e-6 and 128, respectively. We use the cosine scheduler to train the networks.
\vspace{2px}

\noindent\textbf{Evaluation}
To evaluate our CvT+ATS model, we use the same setup as \cite{cvt}.

\subsection{PS-ViT + ATS}
\noindent\textbf{Training}
To fine-tune our adaptive models, we follow the PS-ViT \cite{psvit} training settings. We use the PS-ViT model's pre-trained weights to initialize the backbones of our adaptive network and train it for 30 epochs using AdamW optimizer. The learning rate and batch size are set to 5e-4 and $8\times96$, respectively. We use the cosine scheduler to train the networks.
\vspace{2px}

\noindent\textbf{Evaluation}
To evaluate our CvT+ATS model, we use the same setup as \cite{psvit}.

\subsection{XViT + ATS}
We integrate our ATS module into the stages 3 to 11 of the XViT \cite{xvit} network.

\noindent\textbf{Training}
To train our adaptive model, we follow most of the XViT \cite{xvit} network's training settings. We use the XViT model's pre-trained weights to initialize the backbone of our adaptive network and train it for 10 epochs using SGD optimizer. The learning rate and batch size are set to 1.5e-6 and 64, respectively. We use the cosine scheduler to train the networks.

\noindent\textbf{Evaluation}
To evaluate our XViT+ATS model, we use the same setup as \cite{xvit}.

\subsection{TimeSformer + ATS}
We integrate our ATS module into the stages 3 to 5 of the TimeSformer \cite{bertasius2021space} network.

\noindent\textbf{Training}
To train our adaptive model, we follow most of the TimeSformer \cite{bertasius2021space} network's training settings. We use the TimeSformer model's pre-trained weights to initialize the backbones of our adaptive networks and train it for 5 epochs using SGD optimizer. The learning rate and batch size are set to 5e-6 and 32, respectively. We use the cosine scheduler to train the networks.

\noindent\textbf{Evaluation}
To evaluate our TimeSformer-HR+ATS and TimeSformer-L+ATS models, we use the same setup as \cite{bertasius2021space}.

\subsection{Integrating ATS into a Transformer Block}

\begin{figure*}[t]
\begin{center}
\end{center}
   \includegraphics[width=1.0\linewidth]{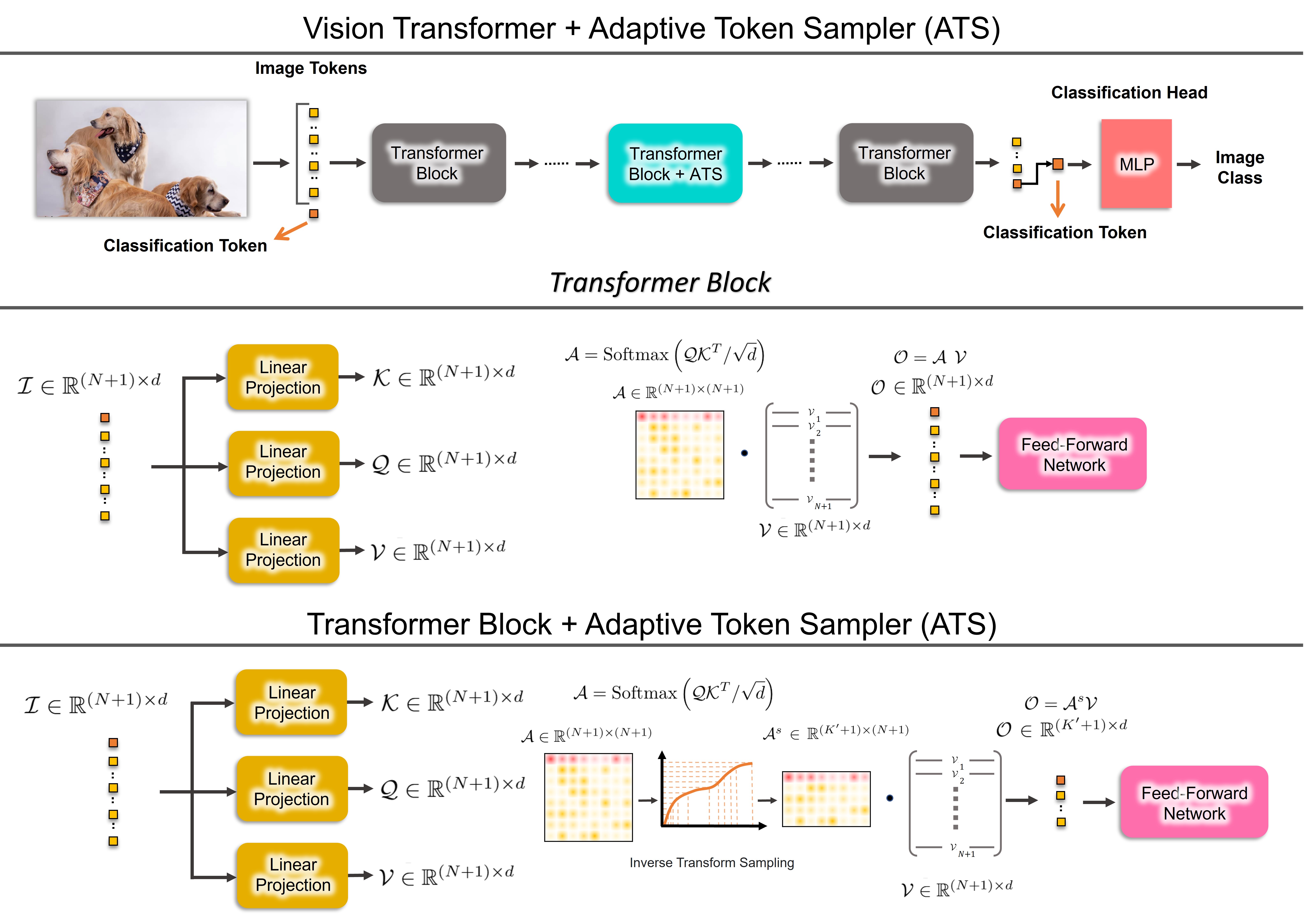}
   \caption{The Adaptive Token Sampler (ATS) can be integrated into the self-attention layer of any transformer block of a vision transformer model (top). 
   The ATS module takes at each stage a set of input tokens $\mathcal{I}$. The first token is considered as the classification token in each block of the vision transformer. The attention matrix $\mathcal{A}$ is then calculated by the dot product of the queries $\mathcal{Q}$ and keys $\mathcal{K}$, scaled by $\sqrt{d}$. Having selected the significant tokens, we then sample the corresponding attention weights (rows of the attention matrix $\mathcal{A}$) to get $\mathcal{A}^s$. Finally, we softly downsample the input tokens $\mathcal{I}$ to output tokens $\mathcal{O}$ using the dot product of $\mathcal{A}^s$ and $\mathcal{V}$. Next, we forward the output tokens $\mathcal{O}$ through a Feed-Forward Network (FFN) to get the output of the transformer block.}
\label{fig:method_supp}
\end{figure*}

Unlike a standard transformer block in vision transformers, we assign a score to each token and use inverse transform sampling to prune the rows of the attention matrix $\mathcal{A}$ to get $\mathcal{A}^s$. Next, we get the output $\mathcal{O}=\mathcal{A}^s\mathcal{V}$ and forward it to the Feed-Forward Network (FFN) of the transformer block.
We visualize the details of our ATS module, which is integrated into a standard self-attention layer in Fig. \ref{fig:method_supp}.   

\section{Ablation}
\subsection{Score Assignment}
\begin{figure}
\RawFloats
\begin{minipage}{\textwidth}
\begin{minipage}[b]{0.5\textwidth}
\centering
\captionof{table}{Comparison of the inverse transform sampling approach with the top-K selection. We finetune and test two different versions of the multi-stage DeiT-S+ATS model: with (1) top-K token selection and (2) inverse transform token sampling. We report the top-1 accuracy of both networks on the ImageNet validation set. For the model with the top-K selection approach, we set $K_n=\round{0.865\times \#{InputTokens}_n}$ where $n$ is the stage index. For example, $K_3=171$ in stage 3.}

\vspace{+0.1in}
\scalebox{0.7}{
    \begin{tabular}{lcc}
    \toprule 
    Method & Top-1 acc & GFLOPs  \\

    \midrule
    Top-K & 78.9 & 2.9 \\
    Inverse Transform Sampling & \textbf{79.7} & 2.9 \\ 
    \bottomrule
    \end{tabular}
}
\label{tab:top-K}
\end{minipage}
\hfill
\begin{minipage}[b]{0.47\textwidth}
    \centering
    \includegraphics[width=\linewidth]{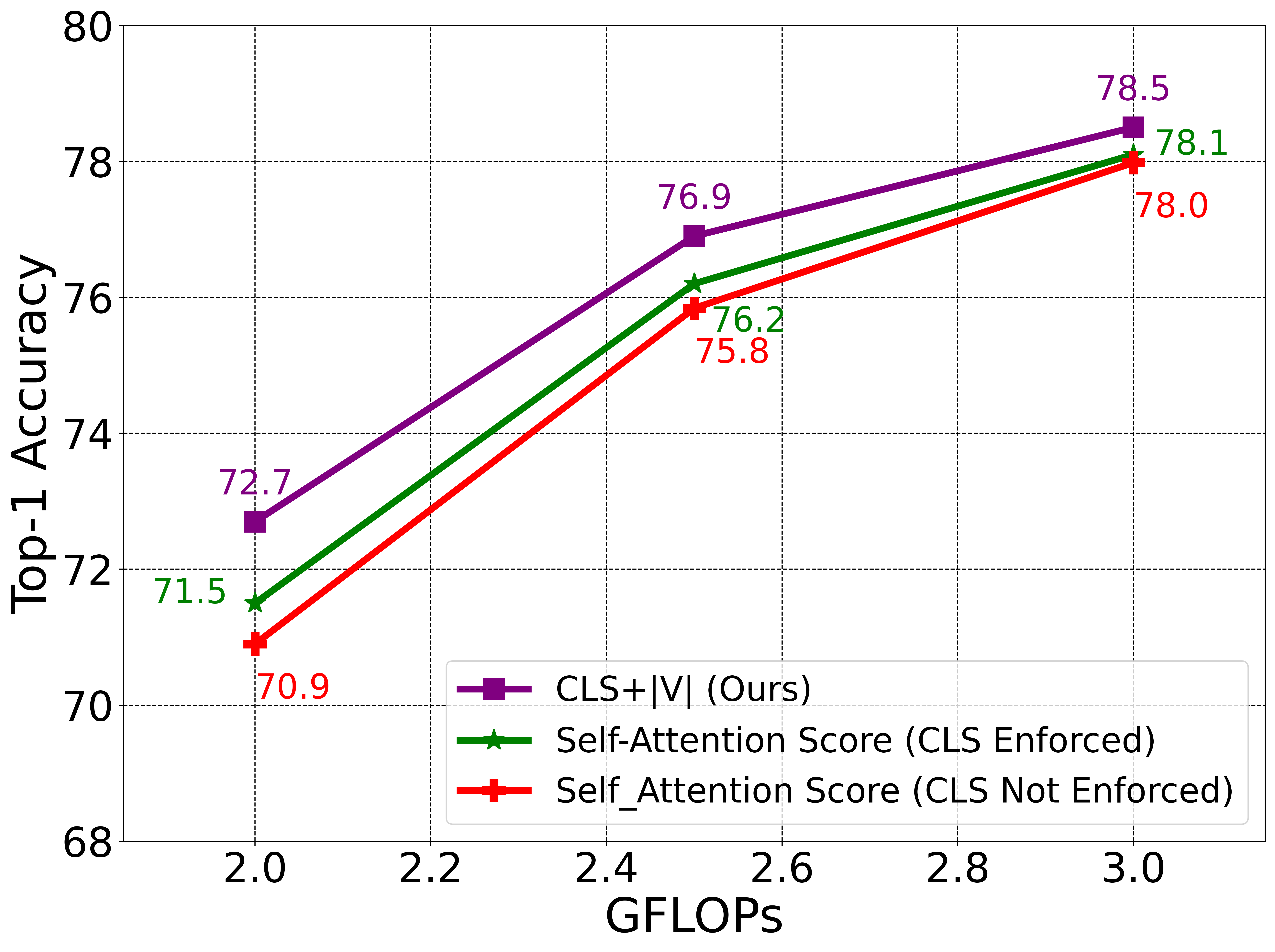}
    \caption{Impact of allowing ATS to discard the classification token on the network's accuracy. The model is a single stage DeiT-S+ATS without finetuning.}
    \label{fig:score_assignment}
\end{minipage}
\end{minipage}
\end{figure}

In the main paper, we analyzed the impact of using different tokens to calculate the significance scores $S$. In all of our experiments, we suggested keeping the classification token since the loss is defined on this token and discarding it may negatively affect the performance. To represent the importance of this token experimentally, we sum over the attention weights of all tokens (rows of the attention matrix) to find the most significant tokens. We show this in Fig. \ref{fig:score_assignment} as Self-Attention Score (CLS Enforced). In contrast to our previous experiments, we allow ATS to remove the classification token when it is of low importance based on the significance scores $S$. We show the results of this experiment in Fig. \ref{fig:score_assignment} as Self-Attention Score (CLS Not Enforced). As it can be seen in Fig. \ref{fig:score_assignment}, discarding the classification token reduces the top-1 accuracy. 

\subsection{Candidate Token Selection}
As mentioned in the main paper, we employ the inverse transform sampling approach to softly downsample input tokens. We investigated this in Section 4 of the paper.
To better analyze it, we also evaluate the performance of our trained multi-stage DeiT-S+ATS model when picking the top K tokens with the highest significance scores $\mathcal{S}$. To this end, we trained our DeiT-S+ATS network with the top-K selection approach and compared it to our DeiT-S+ATS model with the inverse transform sampling method. As it can be seen in Table~\ref{tab:top-K}, our inverse transform sampling approach outperforms the top-K selection with and without training (Fig~5(a) in paper). As discussed earlier, our inverse transform sampling approach does not hardly discard all tokens with lower significance scores and hence provides a more diverse set of tokens for the following layers. This sampling strategy also helps the model to gain a better performance after training, thanks to a more diversified token selection.

\subsection{ATS Placement}
To evaluate the effect of our ATS module's location within a vision transformer model, we add it to different stages of the DeiT-S network and evaluate it on the ImageNet validation set without finetuning the model.  To have a better comparison, we set the average computation costs of all experiments to $3$ GFLOPs. As it can be seen in Table~\ref{tab:ATS_loc}, integrating the ATS module into the first stage of the DeiT-S model results in a poor top-1 accuracy of {$73.1\%$}. On the other hand, integrating the ATS module into stage 3 results in a $78.5\%$ top-1 accuracy. As mentioned before, earlier transformer blocks are more prone to predict noisier attention weights for the classification token. Therefore, integrating our ATS module into the first stage performs worse than incorporating it into the stage 3. Although the attention weights of the stage 6 are less noisy, we have to discard more tokens to reach the desired GFLOPs level of 3. For example in stages 0, 3, and 6, we set $K$ to $130$, $108$, and $56$, respectively. The highest accuracy is obtained when we integrate the ATS module into multiple stages of the DeiT-S model. This is because of the progressive token sampling that occurs in a multi-stage DeiT-S+ATS model. In other words, a multi-stage DeiT-S+ATS network can gradually decrease the GFLOPs by discarding fewer tokens in the earlier stages, while a single-stage DeiT-S+ATS model has to discard more tokens in the earlier stages to reach the same GFLOPs level. We also added the ATS module into all stages, yielding average GFLOPs of 2.6 and $76.9\%$ top-1 accuracy. 

\begin{table}[h]
\centering
\caption{Evaluating the integration of the ATS module into different stages of DeiT-S \cite{deit}.}
\setlength{\tabcolsep}{10pt}
\scalebox{1.0}{
    \begin{tabular}{lccccc}
    \toprule 
    Stage(s) & 0 & 3 & 6 & 3-11 \\
    \midrule
    Top-1 Accuracy & 73.1 & 78.5 & 77.4 & \textbf{79.2}  \\
    \bottomrule
    \end{tabular}
}
\label{tab:ATS_loc}
\end{table}

\subsection{Adding ATS to Models with Other Token Pruning Approaches}
To better evaluate the performance of our adaptive token sampling approach, we also add our module to the state-of-the-art efficient vision transformer EViT-DeiT-S~\cite{evit}. EViT~\cite{evit} introduces a token reorganization method that first identifies the top-K important tokens by computing token attentiveness between the tokens and the classification token and then fuses less informative tokens. Interestingly, our ATS module can also be added to the EViT-DeiT-S model and further decrease the GFLOPs, as shown in Table~\ref{tab:ATS_EViT}. These results demonstrate the superiority of our adaptive token sampling approach compared to static token pruning methods. We integrate our ATS module into stages 4, 5, 7, 8, 10, and 11 of the EViT-DeiT-S backbone and fine-tune them for 10 epochs following our fine-tuning setups on the ImageNet dataset discussed earlier.

\begin{table}[h]
\caption{Evaluating the EViT-DeiT-S~\cite{evit} model's performance when integrating the ATS module into it with $K_n=\round{0.7\times \#{InputTokens}_n}$ where $n$ is the stage index.}
\begin{center}
\scalebox{1.0}{
    \begin{tabular}{lccc}
    \toprule 
    Model & Top-1 acc & GFLOPs \\

    \midrule
    EViT-DeiT-S (30 Epochs)\cite{evit} & 79.5 & 3.0\\
    EViT-DeiT-S (30 Epochs)+ATS & 79.5 & \textbf{2.5} \\
    EViT-DeiT-S (100 Epochs)\cite{evit} & 79.8 & 3.0 \\
    EViT-DeiT-S (100 Epochs)+ATS & 79.8 & \textbf{2.5} \\
    \bottomrule
    \end{tabular}
}
\end{center}

\label{tab:ATS_EViT}
\end{table}

\section{More Visualizations}

\begin{figure*}[htb!]
    \centering
    \includegraphics[width=\linewidth]{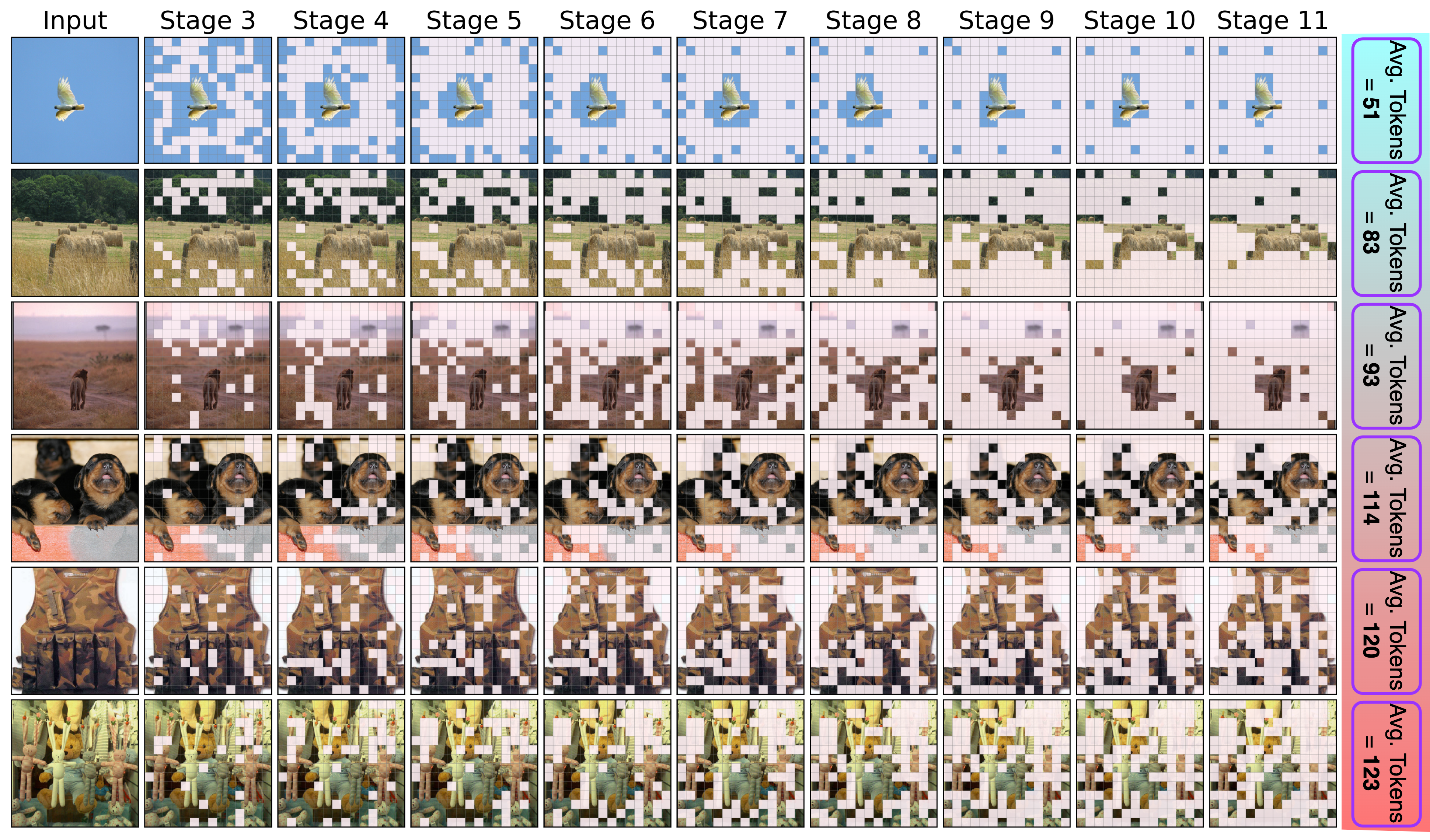} 
    \caption{Visualization of the gradual token sampling procedure in the multi-stage DeiT-S+ATS model. We integrate our ATS module into the stages 3 to 11 of the DeiT-S model. The tokens that are sampled at each stage of the network are shown for images that are ordered by their complexity (from \textcolor{cyan}{low} complexity to \textcolor{red}{high} complexity). We visualize the tokens, which are discarded, as masks over the input images.  
    As it can be seen, a \textcolor{red}{higher} number of tokens are sampled for more cluttered images while a \textcolor{cyan}{lower} number of tokens are required when the images contain less details. Additionally, we can see that the sampled tokens are more focused and less scattered in images with less details.}
    \label{fig:visulization}
\end{figure*}
We show more visual results in Fig. \ref{fig:visulization}. We select several images of the ImageNet validation set with various amounts of detail and complexity.  We visualize the progressive token sampling procedure of our multi-stage DeiT-S+ATS model for the selected images. The number of output tokens of each ATS module in the multi-stage DeiT-S+ATS model is limited by the number of its input tokens, which is 197. Our adaptive model samples a higher number of tokens when the input images are more cluttered. We can also observe that the sampled tokens are more scattered in images with more details compared to more plain images.

\clearpage
%
%
\bibliographystyle{splncs04}
\bibliography{egbib}

\begin{thebibliography}{10}
\providecommand{\url}[1]{\texttt{#1}}
\providecommand{\urlprefix}{URL }
\providecommand{\doi}[1]{https://doi.org/#1}

\bibitem{bertasius2021space}
Bertasius, G., Wang, H., Torresani, L.: Is space-time attention all you need
  for video understanding. In: International Conference on Machine Learning
  (ICML) (2021)

\bibitem{xvit}
Bulat, A., Perez~Rua, J.M., Sudhakaran, S., Martinez, B., Tzimiropoulos, G.:
  Space-time mixing attention for video transformer. In: Advances in Neural
  Information Processing Systems (NeurIPS) (2021)

\bibitem{detr}
Carion, N., Massa, F., Synnaeve, G., Usunier, N., Kirillov, A., Zagoruyko, S.:
  End-to-end object detection with transformers. In: European Conference on
  Computer Vision (ECCV) (2020)

\bibitem{kinetics-600}
Carreira, J., Noland, E., Banki-Horvath, A., Hillier, C., Zisserman, A.: A
  short note about kinetics-600. In: arXiv preprint arXiv:1808.01340v1 (2018)

\bibitem{chen2021crossvit}
Chen, C.F., Fan, Q., Panda, R.: Crossvit: Cross-attention multi-scale vision
  transformer for image classification. In: IEEE/CVF International Conference
  on Computer Vision (ICCV) (2021)

\bibitem{maskformer}
Cheng, B., Schwing, A.G., Kirillov, A.: Per-pixel classification is not all you
  need for semantic segmentation. In: Advances in Neural Information Processing
  Systems (NeurIPS) (2021)

\bibitem{sparsetransformer}
Child, R., Gray, S., Radford, A., Sutskever, I.: Generating long sequences with
  sparse transformers. In: arXiv preprint arXiv:1904.10509 (2019)

\bibitem{chu2021cpvt}
Chu, X., Tian, Z., Zhang, B., Wang, X., Wei, X., Xia, H., Shen, C.: Conditional
  positional encodings for vision transformers. In: arXiv preprint
  arXiv:2102.10882 (2021)

\bibitem{deng2009imagenet}
Deng, J., Dong, W., Socher, R., Li, L.J., Li, K., Fei-Fei, L.: Imagenet: A
  large-scale hierarchical image database. In: IEEE/CVF Conference on Computer
  Vision and Pattern Recognition (CVPR) (2009)

\bibitem{diba2018spatio}
Diba, A., Fayyaz, M., Sharma, V., Arzani, M.M., Yousefzadeh, R., Gall, J.,
  Van~Gool, L.: Spatio-temporal channel correlation networks for action
  classification. In: European Conference on Computer Vision (ECCV) (2018)

\bibitem{diba2020large}
Diba, A., Fayyaz, M., Sharma, V., Paluri, M., Gall, J., Stiefelhagen, R., Gool,
  L.V.: Large scale holistic video understanding. In: European Conference on
  Computer Vision (2020)

\bibitem{dynamonet}
Diba, A., Sharma, V., Gool, L.V., Stiefelhagen, R.: Dynamonet: Dynamic action
  and motion network. In: IEEE/CVF International Conference on Computer Vision
  (ICCV) (2019)

\bibitem{vit}
Dosovitskiy, A., Beyer, L., Kolesnikov, A., Weissenborn, D., Zhai, X.,
  Unterthiner, T., Dehghani, M., Minderer, M., Heigold, G., Gelly, S.,
  Uszkoreit, J., Houlsby, N.: An image is worth 16x16 words: Transformers for
  image recognition at scale. In: International Conference on Learning
  Representations (ICLR) (2021)

\bibitem{fan2021multiscale}
Fan, H., Xiong, B., Mangalam, K., Li, Y., Yan, Z., Malik, J., Feichtenhofer,
  C.: Multiscale vision transformers. In: IEEE/CVF International Conference on
  Computer Vision (ICCV) (2021)

\bibitem{fan2019blvnet}
Fan, Q., Chen, C.F.R., Kuehne, H., Pistoia, M., Cox, D.: {More Is Less:
  Learning Efficient Video Representations by Temporal Aggregation Modules}.
  In: Advances in Neural Information Processing Systems (NeurIPS) (2019)

\bibitem{atfr}
Fayyaz, M., Bahrami, E., Diba, A., Noroozi, M., Adeli, E., Van~Gool, L., Gall,
  J.: 3d cnns with adaptive temporal feature resolutions. In: IEEE/CVF
  Conference on Computer Vision and Pattern Recognition (CVPR) (2021)

\bibitem{feichtenhofer2020x3d}
Feichtenhofer, C.: X3d: Expanding architectures for efficient video
  recognition. In: IEEE/CVF Conference on Computer Vision and Pattern
  Recognition (CVPR) (2020)

\bibitem{feichtenhofer2019slowfast}
Feichtenhofer, C., Fan, H., Malik, J., He, K.: Slowfast networks for video
  recognition. In: IEEE/CVF international conference on computer vision (ICCV)
  (2019)

\bibitem{gong2014compressing}
Gong, Y., Liu, L., Yang, M., Bourdev, L.: Compressing deep convolutional
  networks using vector quantization. In: arXiv preprint arXiv:1412.6115 (2014)

\bibitem{powerbert}
Goyal, S., Choudhury, A.R., Raje, S.M., Chakaravarthy, V.T., Sabharwal, Y.,
  Verma, A.: Power-bert: Accelerating bert inference via progressive
  word-vector elimination. In: International Conference on Machine Learning
  (ICML) (2020)

\bibitem{startransformer}
Guo, Q., Qiu, X., Liu, P., Shao, Y., Xue, X., Zhang, Z.: Star-transformer. In:
  arXiv preprint arXiv:1902.09113 (2019)

\bibitem{han2021transformer}
Han, K., Xiao, A., Wu, E., Guo, J., Xu, C., Wang, Y.: Transformer in
  transformer. In: Advances in Neural Information Processing Systems (NeurIPS)
  (2021)

\bibitem{he2015deep}
He, K., Zhang, X., Ren, S., Sun, J.: Deep residual learning for image
  recognition. In: IEEE/CVF Conference on Computer Vision and Pattern
  Recognition (CVPR) (2016)

\bibitem{channel_pruning}
He, Y., Zhang, X., Sun, J.: Channel pruning for accelerating very deep neural
  networks. In: IEEE/CVF International Conference on Computer Vision (ICCV)
  (2017)

\bibitem{hinton2015distilling}
Hinton, G., Vinyals, O., Dean, J.: Distilling the knowledge in a neural
  network. In: NIPS Deep Learning and Representation Learning Workshop (2015)

\bibitem{mobilenets}
Howard, A.G., Zhu, M., Chen, B., Kalenichenko, D., Wang, W., Weyand, T.,
  Andreetto, M., Adam, H.: Mobilenets: Efficient convolutional neural networks
  for mobile vision applications. In: arXiv preprint arXiv:1704.04861 (2017)

\bibitem{low_rank_factorization_2}
Jaderberg, M., Vedaldi, A., Zisserman, A.: Speeding up convolutional neural
  networks with low rank expansions. In: arXiv preprint arXiv:1405.3866 (2014)

\bibitem{scaling-transformers}
Jaszczur, S., Chowdhery, A., Mohiuddin, A., Kaiser, L., Gajewski, W.,
  Michalewski, H., Kanerva, J.: Sparse is enough in scaling transformers. In:
  Advances in Neural Information Processing Systems (NeurIPS) (2021)

\bibitem{jiang2019stm}
Jiang, B., Wang, M., Gan, W., Wu, W., Yan, J.: Stm: Spatiotemporal and motion
  encoding for action recognition. In: IEEE/CVF International Conference on
  Computer Vision (ICCV) (2019)

\bibitem{jiang2021token}
Jiang, Z., Hou, Q., Yuan, L., Zhou, D., Jin, X., Wang, A., Feng, J.: Token
  labeling: Training a 85.5\% top-1 accuracy vision transformer with 56m
  parameters on imagenet. In: arXiv preprint arXiv:2104.10858v2 (2021)

\bibitem{token_labeling}
Jiang, Z., Hou, Q., Yuan, L., Zhou, D., Shi, Y., Jin, X., Wang, A., Feng, J.:
  All tokens matter: Token labeling for training better vision transformers.
  In: Advances in Neural Information Processing Systems (NeurIPS) (2021)

\bibitem{tinybert}
Jiao, X., Yin, Y., Shang, L., Jiang, X., Chen, X., Li, L., Wang, F., Liu, Q.:
  Tinybert: Distilling bert for natural language understanding. In: arXiv
  preprint arXiv:1909.10351 (2020)

\bibitem{kinetics-400}
Kay, W., Carreira, J., Simonyan, K., Zhang, B., Hillier, C., Vijayanarasimhan,
  S., Viola, F., Green, T., Back, T., Natsev, A., Suleyman, M., Zisserman, A.:
  The kinetics human action video dataset. In: arXiv preprint arXiv:1705.06950
  (2017)

\bibitem{krizhevsky2014weird}
Krizhevsky, A.: One weird trick for parallelizing convolutional neural
  networks. In: ArXiv preprint arXiv:1404.5997 (2014)

\bibitem{li2020tea}
Li, Y., Ji, B., Shi, X., Zhang, J., Kang, B., Wang, L.: Tea: Temporal
  excitation and aggregation for action recognition. In: IEEE/CVF Conference on
  Computer Vision and Pattern Recognition (CVPR) (2020)

\bibitem{evit}
Liang, Y., Ge, C., Tong, Z., Song, Y., Wang, J., Xie, P.: Not all patches are
  what you need: Expediting vision transformers via token reorganizations. In:
  International Conference on Learning Representations (ICLR) (2022)

\bibitem{lin2019tsm}
Lin, J., Gan, C., Han, S.: Tsm: Temporal shift module for efficient video
  understanding. In: IEEE/CVF International Conference on Computer Vision
  (ICCV) (2019)

\bibitem{metadistiller}
Liu, B., Rao, Y., Lu, J., Zhou, J., jui Hsieh, C.: Metadistiller: Network
  self-boosting via meta-learned top-down distillation. In: European Conference
  on Computer Vision (ECCV) (2020)

\bibitem{swin}
Liu, Z., Lin, Y., Cao, Y., Hu, H., Wei, Y., Zhang, Z., Lin, S., Guo, B.: Swin
  transformer: Hierarchical vision transformer using shifted windows. In:
  Proceedings of the IEEE/CVF International Conference on Computer Vision
  (ICCV) (2021)

\bibitem{token-pooling}
Marin, D., Chang, J.H.R., Ranjan, A., Prabhu, A.K., Rastegari, M., Tuzel, O.:
  Token pooling in vision transformers. arXiv preprint arXiv:2110.03860  (2021)

\bibitem{IARED}
Pan, B., Panda, R., Jiang, Y., Wang, Z., Feris, R., Oliva, A.: {IA}-{RED}$^2$:
  Interpretability-aware redundancy reduction for vision transformers. In:
  Advances in Neural Information Processing Systems (NeurIPS) (2021)

\bibitem{HVT}
Pan, Z., Zhuang, B., Liu, J., He, H., Cai, J.: Scalable vision transformers
  with hierarchical pooling. In: IEEE/CVF International Conference on Computer
  Vision (ICCV) (2021)

\bibitem{qiu2019learning}
Qiu, Z., Yao, T., Ngo, C.W., Tian, X., Mei, T.: Learning spatio-temporal
  representation with local and global diffusion. In: IEEE/CVF Conference on
  Computer Vision and Pattern Recognition (CVPR) (2019)

\bibitem{radosavovic2020designing}
Radosavovic, I., Kosaraju, R.P., Girshick, R., He, K., Doll{\'a}r, P.:
  Designing network design spaces. In: IEEE/CVF Conference on Computer Vision
  and Pattern Recognition (CVPR) (2020)

\bibitem{runtime_net_prunning}
Rao, Y., Lu, J., Lin, J., Zhou, J.: Runtime network routing for efficient image
  classification. In: IEEE Transactions on Pattern Analysis and Machine
  Intelligence. vol.~41, pp. 2291--2304 (2019)

\bibitem{dynamicvit}
Rao, Y., Zhao, W., Liu, B., Lu, J., Zhou, J., Hsieh, C.J.: Dynamicvit:
  Efficient vision transformers with dynamic token sparsification. In: Advances
  in Neural Information Processing Systems (NeurIPS) (2021)

\bibitem{global_filter_net}
Rao, Y., Zhao, W., Zhu, Z., Lu, J., Zhou, J.: Global filter networks for image
  classification. In: Advances in Neural Information Processing Systems
  (NeurIPS) (2021)

\bibitem{routingtransformer}
Roy, A., Saffar, M., Vaswani, A., Grangier, D.: Efficient content-based sparse
  attention with routing transformers. In: Transactions of the Association for
  Computational Linguistics. vol.~9, pp. 53--68 (2021)

\bibitem{tokenlearner}
Ryoo, M.S., Piergiovanni, A., Arnab, A., Dehghani, M., Angelova, A.:
  Tokenlearner: What can 8 learned tokens do for images and videos? arXiv
  preprint arXiv:2106.11297  (2021)

\bibitem{simonyan2015deep}
Simonyan, K., Zisserman, A.: Very deep convolutional networks for large-scale
  image recognition. arXiv preprint arXiv:1409.1556  (2015)

\bibitem{adaptive-span}
Sukhbaatar, S., Grave, E., Bojanowski, P., Joulin, A.: Adaptive attention span
  in transformers. In: ACL (2019)

\bibitem{efficient_net}
Tan, M., Le, Q.: {E}fficient{N}et: Rethinking model scaling for convolutional
  neural networks. In: International Conference on Machine Learning (ICML)
  (2019)

\bibitem{deit}
Touvron, H., Cord, M., Douze, M., Massa, F., Sablayrolles, A., Jegou, H.:
  Training data-efficient image transformers and distillation through
  attention. In: International Conference on Machine Learning (ICML) (2021)

\bibitem{c3d}
Tran, D., Bourdev, L., Fergus, R., Torresani, L., Paluri, M.: Learning
  spatiotemporal features with 3d convolutional networks. In: IEEE
  International Conference on Computer Vision (ICCV) (2015)

\bibitem{tran2019video}
Tran, D., Wang, H., Torresani, L., Feiszli, M.: Video classification with
  channel-separated convolutional networks. In: IEEE/CVF International
  Conference on Computer Vision (ICCV) (2019)

\bibitem{res3d}
Tran, D., Wang, H., Torresani, L., Ray, J., LeCun, Y., Paluri, M.: A closer
  look at spatiotemporal convolutions for action recognition. In: IEEE
  Conference on Computer Vision and Pattern Recognition (CVPR) (2018)

\bibitem{attention_is_all_you_need}
Vaswani, A., Shazeer, N., Parmar, N., Uszkoreit, J., Jones, L., Gomez, A.N.,
  Kaiser, L.u., Polosukhin, I.: Attention is all you need. In: Advances in
  Neural Information Processing Systems (NeuRIPS) (2017)

\bibitem{wang2020video}
Wang, H., Tran, D., Torresani, L., Feiszli, M.: Video modeling with correlation
  networks. In: IEEE/CVF Conference on Computer Vision and Pattern Recognition
  (CVPR) (2020)

\bibitem{haq}
Wang, K., Liu, Z., Lin, Y., Lin, J., Han, S.: Haq: Hardware-aware automated
  quantization with mixed precision. In: IEEE Conference on Computer Vision and
  Pattern Recognition (CVPR) (2019)

\bibitem{wang2021pvt}
Wang, W., Xie, E., Li, X., Fan, D.P., Song, K., Liang, D., Lu, T., Luo, P.,
  Shao, L.: Pyramid vision transformer: A versatile backbone for dense
  prediction without convolutions. In: IEEE/CVF International Conference on
  Computer Vision (ICCV) (2021)

\bibitem{attentionnas}
Wang, X., Xiong, X., Neumann, M., Piergiovanni, A.J., Ryoo, M.S., Angelova, A.,
  Kitani, K.M., Hua, W.: Attentionnas: Spatiotemporal attention cell search for
  video classification. In: European Conference on Computer Vision (ECCV)
  (2020)

\bibitem{wang2018non}
Wang, X., Girshick, R., Gupta, A., He, K.: Non-local neural networks. In:
  IEEE/CVF Conference on Computer Vision and Pattern Recognition (CVPR) (2018)

\bibitem{cvt}
Wu, H., Xiao, B., Codella, N., Liu, M., Dai, X., Yuan, L., Zhang, L.: Cvt:
  Introducing convolutions to vision transformers. In: IEEE/CVF International
  Conference on Computer Vision (ICCV) (2021)

\bibitem{xu2021coat}
Xu, W., Xu, Y., Chang, T., Tu, Z.: Co-scale conv-attentional image
  transformers. In: arXiv preprint arXiv:2104.06399 (2021)

\bibitem{low_rank_factorization_1}
Yu, X., Liu, T., Wang, X., Tao, D.: On compressing deep models by low rank and
  sparse decomposition. In: IEEE/CVF Conference on Computer Vision and Pattern
  Recognition (CVPR) (2017)

\bibitem{pointr}
Yu, X., Rao, Y., Wang, Z., Liu, Z., Lu, J., Zhou, J.: Pointr: Diverse point
  cloud completion with geometry-aware transformers. In: IEEE/CVF International
  Conference on Computer Vision (ICCV) (2021)

\bibitem{yuan2021t2t}
Yuan, L., Chen, Y., Wang, T., Yu, W., Shi, Y., Jiang, Z., Tay, F.E., Feng, J.,
  Yan, S.: Tokens-to-token vit: Training vision transformers from scratch on
  imagenet. In: arXiv preprint arXiv:2101.11986 (2021)

\bibitem{psvit}
Yue, X., Sun, S., Kuang, Z., Wei, M., Torr, P., Zhang, W., Lin, D.: Vision
  transformer with progressive sampling. In: IEEE/CVF International Conference
  on Computer Vision (ICCV) (2021)

\bibitem{pointtransformer}
Zhao, H., Jiang, L., Jia, J., Torr, P., Koltun, V.: Point transformer. In:
  IEEE/CVF International Conference on Computer Vision (ICCV) (2021)

\bibitem{setr}
Zheng, S., Lu, J., Zhao, H., Zhu, X., Luo, Z., Wang, Y., Fu, Y., Feng, J.,
  Xiang, T., Torr, P.H., Zhang, L.: Rethinking semantic segmentation from a
  sequence-to-sequence perspective with transformers. In: IEEE/CVF Conference
  on Computer Vision and Pattern Recognition (CVPR) (2021)

\bibitem{deepvit}
Zhou, D., Kang, B., Jin, X., Yang, L., Lian, X., Jiang, Z., Hou, Q., Feng, J.:
  Deepvit: Towards deeper vision transformer. arXiv preprint arXiv:2103.11886
  (2021)

\end{thebibliography}
\end{document}